%% file: main.tex
\pdfoutput=1

\documentclass[11pt]{article}

\usepackage[final]{acl}

\usepackage{times}
\usepackage{latexsym}

\usepackage[T1]{fontenc}

\usepackage[utf8]{inputenc}

\usepackage{microtype}

\usepackage{inconsolata}

\usepackage{graphicx} 
\usepackage{booktabs} 
\usepackage{makecell}  
\usepackage{subfig} 
\usepackage{amsmath} 
\usepackage{multirow} 
\usepackage{colortbl} 
\usepackage{tcolorbox} 
\usepackage{enumitem} 

\newcommand{\ENDow}{{\textsc{ENDow}}}

\title{Measuring the Effect of Transcription Noise on\\Downstream Language Understanding Tasks}

\author{{\bf Ori Shapira}, {\bf Shlomo E. Chazan}, \and {\bf Amir DN Cohen} \\
        OriginAI \\ \texttt{oris@originai.co}}

\begin{document}
\maketitle

\input{00_abstract}

\input{01_introduction}

\input{02_related_work}
\input{03_framework}

\input{04_setup}

\input{05_results}
\input{06_conclusion}

\input{07_limitations}


\bibliography{bibliography}

\newpage
\input{0a_appendix}

\end{document}

%% file: 00_abstract.tex
\begin{abstract}

With the increasing prevalence of recorded human speech, spoken language understanding (SLU) is essential for its efficient processing. In order to process the speech, it is commonly transcribed using automatic speech recognition technology. This speech-to-text transition introduces errors into the transcripts, which subsequently propagate to downstream NLP tasks, such as dialogue summarization. While it is known that transcript noise affects downstream tasks, a systematic approach to analyzing its effects across different noise severities and types has not been addressed. We propose a configurable framework for assessing task models in diverse noisy settings, and for examining the impact of transcript-cleaning techniques. The framework facilitates the investigation of task model behavior, which can in turn support the development of effective SLU solutions. We exemplify the utility of our framework on three SLU tasks and four task models, offering insights regarding the effect of transcript noise on tasks in general and models in particular. For instance, we find that task models can tolerate a certain level of noise, and are affected differently by the types of errors in the transcript.\footnote{Code: \url{https://github.com/OriShapira/ENDow}}

\end{abstract}

%% file: 01_introduction.tex
\section{Introduction}
\label{sec_introduction}

Human speech is captured by microphones constantly. Dialogues or utterances are recorded at online meetings, for creating content, and for being aided by virtual assistants or service providers. Many of these recordings inevitably require automated processing, for which the common approach is to run automatic speech recognition (ASR) systems that convert audio to transcribed text. The produced transcript is then handled
with spoken language understanding (SLU) technology. 

Considerable effort is invested in developing ASR systems that can overcome environmental sounds, vague speech and phenomena of spoken language,
in order to produce transcripts that are as faithful to the speech (``clean'') as possible \citep{iwamoto2022artifacts, Prabhavalkar2023surveyASR}. In turn, the text processing step can be performed more effectively. Simply put, the mistakes (``noise'') produced in the speech-to-text stage propagate to downstream tasks in the text processing stage \citep{kubis-etal-2023-back, feng2022asrglue}.

\input{figures/example}

Eventual SLU tasks are abundant, from traditional dialog act classification \citep{shriberg-etal-2004-icsi} to summarization \citep{Waibel1998summ} and even neurological assessment of speakers \citep{Roshanzamir2021Alzheimer}. 
Indeed, over the years studies have noticed that noisy transcripts burden NLP models, and actions are consequently taken to work around or mitigate the noise (surveyed in Section \ref{sec_related_work}). Furthermore, different downstream tasks are not alike in how they respond to the amount and types of errors in transcripts. Some are highly vulnerable to errors, while others may tolerate more noise, or specific types of noise, depending on a task's requirements (as demonstrated in our analyses in Section \ref{sec_results}).
\autoref{fig_example} shows an utterance transcribed with varying levels of error severity, causing unpredictable behavior in downstream understanding tasks. 
Importantly, the standard word error rate (WER) metric, that measures the amount of errors in generated transcripts, does not capture discrepancies in types of noise, and cannot forecast results on downstream tasks \citep{wang2003indicator}.


Drawing upon lessons from previous research on SLU, in this work we propose a 
framework for systematically analyzing the \textit{\textbf{e}}ffect of transcription \textit{\textbf{n}}oise on a \textit{\textbf{dow}}nstream task (\ENDow; Section \ref{sec_framework}). Our first-of-its-kind framework examines task model behavior under varying noise intensities and types, providing quantitative metrics and facilitating qualitative analyses.
It determines acceptable noise levels for a downstream task, identifies effective transcript-cleaning techniques, and supports planning and implementation of SLU solutions.
Previous studies have examined aspects of \ENDow{}, but always within the scope of a specific task or use case. We suggest that, especially in the era of generalized models and benchmarks, there is a need for a versatile framework that consistently analyzes and compares SLU solutions.
The components in the framework's pipeline, such as the ASR system or task model, are flexibly configured to perform controlled examinations.


Given an SLU dataset, the framework prepares audio files, with varying levels of acoustic distortion, which are then transcribed by an ASR system, producing transcript sets with increasing levels of transcription noise. A method of transcript cleaning then adjusts noise \textit{types}, generating additional transcript versions. Finally, a downstream task model is applied, allowing comparison and analysis across the transcript versions.
Notably, beyond its configurable pipeline, the framework supports any task dataset, including non-spoken language datasets, greatly expanding the scope for assessing \ENDow{}.

We exemplify the use of our framework (Section \ref{sec_setup}), and perform an extensive analysis (Section \ref{sec_results}) across three SLU tasks, with seven intensities of noise, seven cleaning techniques, and four LLM task models. Specifically, we focus on 
summarization
\citep{zhong-etal-2021-qmsum}, 
question-answering
\citep{wu-etal-2022-qaconv}, and dialog-act classification
\citep{shriberg-etal-2004-icsi}, all from existing SLU datasets. 

The results of our diversified experiments yield many insights regarding the level of noise that is acceptable for the downstream tasks, and the impact of the \textit{type} of errors in the transcripts. For example, we observe that named entities are usually the most important term-types for dealing with the tasks, while, surprisingly, verbs are seemingly not as essential. Some findings are unique to specific tasks and models, while others are more consistent. For instance, it is apparent in our experiments that there is a certain amount of noise from which it is not worth the trouble of reducing it. However, that intensity fluctuates with respect to the task, the model, and the type of noise.
The framework also reveals phenomena that occur at particular noise levels, as well as gradual changes as noise increases. For example, we find that GPT-4o-mini \citep{openai2024gpt4ocard} outperforms other models on summarization when transcript noise is low, but the other models overtake GPT as noise increases.
Such findings are valuable for identifying commonalities and differences among various SLU configurations, helping to prioritize efforts for achieving satisfactory outcomes on a downstream task.

%% file: figures/example.tex
\begin{figure}[t]
    \centering
    \includegraphics[width=0.7\linewidth, trim=172 234 413 149, clip]{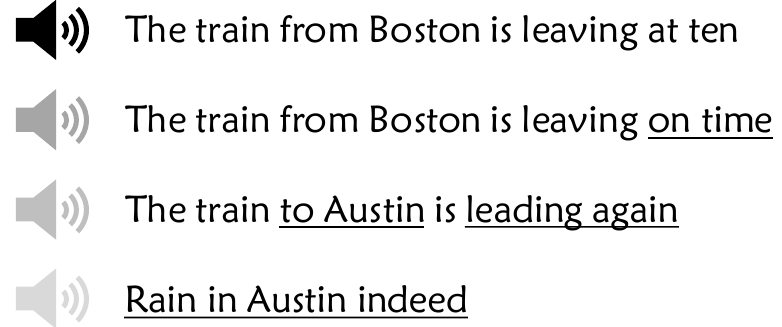}
    \caption{Speech can be transcribed with varying levels of error severity,
    which effects the results of downstream language understanding tasks. For example, summarizing a transcript with variations of the utterance above might produce differing outcomes. The top version is the reference, and the following are \underline{marked} with errors.}
    \label{fig_example}
\end{figure}

%% file: 02_related_work.tex
\input{figures/framework}

\section{Background and Related Work}
\label{sec_related_work}

Spoken language understanding \citep[SLU;][]{wang2005slu} commonly refers to a set of applicative tasks performed on speech \citep{feng2022asrglue, shon-etal-2023-slue}. A prevalent approach for SLU is to transcribe speech with ASR systems, and to process the text with Natural Language Understanding (NLU) techniques that extract meaning from it \citep{tur2011slu}.

Spoken language does not usually obey standard syntactic rules and contains disfluencies such as repairs and hesitations \citep{wang2005slu}. Moreover, when recording speech, the distance of a microphone, the clarity of speech, and environmental sounds add challenging hurdles for an ASR system, that hence produce transcripts that are unfaithful to the spoken words. The discrepancies between the reference (gold) transcript and the automatically produced transcript, a.k.a. ``noise'', is most commonly measured with \textit{word error rate} (WER). This metric measures the percentage of words in the ASR-transcript that were wrongly inserted, substituted and deleted, with respect to the reference transcript, i.e., a form of word edit distance.

To decrease the WER scores or improve subsequent results on downstream tasks, one line of work focuses on correcting ASR-generated transcripts, e.g., by correcting spelling \citep{guo2019cleaning, dutta2022errorcorrectionasrusing}, disfluencies \citep{Stouten2006cleaning}, punctuation \citep{di-gangi-etal-2019-robust}, or mistakes in general \citep{leng-etal-2021-fastcorrect-2, guo2023cleaning}.
However, no cleaning method is flawless, and the errors in the transcript propagate on to the NLU stage \citep{Errattahi2018correction}. 
Additionally, studies \citep{gopalakrishnan2020dialog, kim2021robust} argue that NLU models are mainly trained on written language and are not robust for spoken language, let alone for erroneous spoken language.
Therefore, some works train proprietary models for downstream tasks with noisy transcripts to improve results, e.g., for machine translation, intent classification, question answering, and more \citep{fang2020phoneme, cui2021approachimproverobustnessnlp, liu-etal-2021-robustness, feng22c_interspeech, jung-etal-2024-interventional}.

Another track of research \textit{analyzes} the influence of transcript noise on tasks such as summarization \citep{Szaszak2016noiseSumm, Tundik2019noiseSLU, Chowdhury2024noiseeffect}, question-answering \citep{lee2018spokenSquad, You2021noiseQA} and classification \citep{shon2022slue, Pentland2023socialscience}, mainly by comparing results with and without transcript noise in the input.
To expand this analysis, there are works that assess downstream results at \textit{several} levels or types of noise \citep{zechner-waibel-2000-minimizing, Agarwal2007noise, gopalakrishnan2020dialog, feng2022asrglue, shon-etal-2023-slue, Li2024dementia}. Some add noise synthetically to reference transcripts, while a few prepare recordings with impaired clarity \citep{feng2022asrglue} or use back-transcription \citep{kubis-etal-2023-back}.
Controlling the types of noise reveals their effect \citep{balagopalan-etal-2020-impact, min-etal-2021-evaluating}, working around the limitations of the WER metric, that does not account for the type of words or their importance to the task \citep{wang2003indicator}.
User studies were similarly set up in order to assess how well \textit{humans} conduct tasks on noisy transcripts \citep{stark2000speechretrieval, sanders2002effect, munteanu2006acceptable, favre2013performance}.
They often find at which WER score the ability of users to consistently complete tasks starts to deter.


The studies described above aim to analyze the effect of transcription noise on downstream tasks, however each concentrates on a specific setting and employs different practices. Our proposed framework generalizes a method for conducting such assessments, allowing systematic examination of SLU pipelines.
Furthermore, although there are few works \citep{li2023gptSLU, zhu-etal-2024-zero} that analyze the ability of GPT-family LLMs to handle short transcribed texts on SLU tasks \citep[e.g., ASR-GLUE;][]{feng2022asrglue}, ours is the first, to the best of our knowledge, to assess the performance of several recent LLMs on full dialogues of spoken language. Regardless, our framework is robust to any form of dialogue, NLU task, and task model.

%% file: figures/framework.tex
\begin{figure*}[ht]
    \centering
    \includegraphics[width=\textwidth, trim=79 280 93 123, clip]{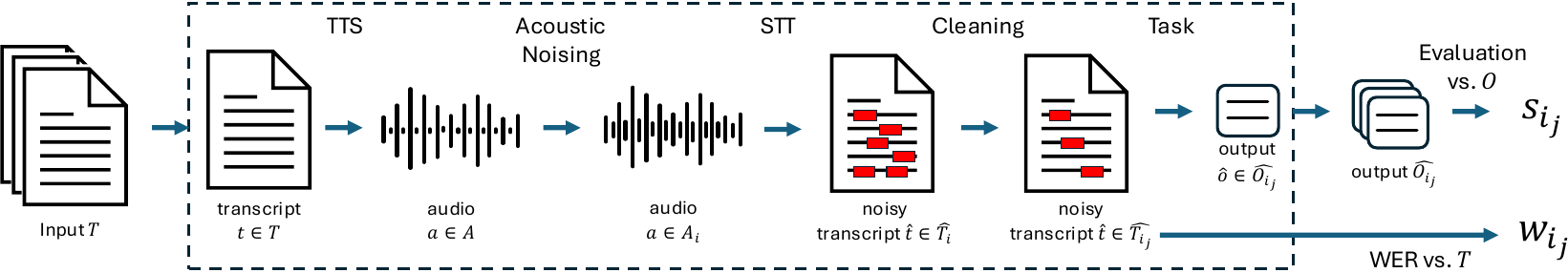}
    \caption{The pipeline of the \ENDow{} framework 
    which yields a downstream task score and a WER score of the transcript set input to the task. The pipeline is executed for several severeties of noising and types of cleaning techniques. 
    Resulting scores are plotted on a graph for the analyses, as in, e.g., \autoref{fig_cleaning_graphs}.}
    \label{fig_framework}
\end{figure*}

%% file: 03_framework.tex
\section{A Framework for Measuring \ENDow{}}
\label{sec_framework}

With the purpose of systematically analyzing SLU pipelines,
our framework's objective is to describe the behavior of downstream tasks as a function of the noise score (e.g., WER, which we use throughout the paper, but any transcription noise metric can be applied) and the type of noise in transcripts. 

The \textbf{input} to the framework is an SLU dataset $D = (T, O)$, where $T$ is a set of reference transcripts and $O$ are the respective expected outcomes. For example, a set of meetings and their respective summaries, for the task of meeting summarization.

The framework consists of a pipeline (illustrated in \autoref{fig_framework}) which includes a text-to-speech (TTS) model to generate audio files for $T$; the acoustic noising method and intensity to apply on the audio; an ASR system for audio transcription; the transcript cleaning technique; the downstream task model; and the evaluation metrics for the task. The components in the pipeline are flexibly set according to the use-case being analyzed.

The framework \textbf{output}s a report on the behavior of the SLU pipeline at the different noise levels and with the cleaning techniques assessed (\S{\ref{sec_framework_analysis}}).





\subsection{Preparing Transcripts with Varying Noise}
\label{sec_framework_noise}

\paragraph{Creating initial audio files.}
Audio files are first created for the input transcripts, in case the SLU dataset lacks them (or when using a non-SLU dataset), or to begin the analysis with clean audio\footnote{That is, with clear speech, without background noise or overlapping speakers.} for greater control over the subsequent noising process.
The TTS system is executed on each input (transcript) in dataset $D$, resulting in the corresponding set of audio files $A$.

\paragraph{Adding noise to audio files.}
Given the audio files $A$, each is acoustically impaired at $k$ levels to increase transcription difficulty, preferably under realistic acoustic conditions.
To that end, reverberation (i.e., sound reflection, like echoing) is applied, and background sounds are added with increasing intensity (signal-to-noise ratio) \citep{wang2018speechsep}.
This stage yields a collection of audio sets $\{A_i\}_{i=1}^k$ (and we define $A_0 = A$), where the severity of impairment increases as $i$ increases.

\paragraph{Transcribing audio files.}
The ASR model is then executed on the audio files in sets $\{A_i\}_{i=0}^k$, resulting in respective transcripts $\{\widehat{T_i}\}_{i=0}^k$. Overall, there are $k+2$ sets of transcripts for dataset $D$: the $k+1$ ASR-generated sets and reference set $T$. It is expected that as $i$ increases, $\widehat{T_i}$ will have a higher WER score (more errors) with respect to $T$.

\paragraph{Cleaning transcripts.}
Each non-reference transcript (in all sets $\widehat{T_i}$) is partially repaired using one of $m$ cleaning techniques. This culminates in sets $\{\{\widehat{T_{i_j}}\}_{j=1}^m\}_{i=0}^k$, and $\widehat{T_{i_0}} = \widehat{T_i}$ (when no cleaning is performed on $\widehat{T_i}$), encompassing $(k+1)*(m+1)$ different levels and types of transcript noise.

\subsection{Executing the Downstream Task}
\label{sec_framework_task}
Next, the task model is executed on each of the transcripts in the prepared transcript sets, producing the respective predicted outputs $\{\{\widehat{O_{i_j}}\}_{j=0}^m\}_{i=0}^k$, and $\widehat{O}$ for the reference transcripts $T$. The predicted outputs in each set $\widehat{O_{i_j}}$ and $\widehat{O}$ are then evaluated against the respective expected outcomes in $O$. 
Finally, this process culminates with the overall score of each dataset variant $\{\{s_{i_j}\}_{j=0}^m\}_{i=0}^k$ and $s$.\footnote{To clarify, $s$ is the score obtained on reference transcripts $T$, portraying a standard execution of the SLU task on input dataset $D$. Score $s_{i_j}$ is for one of the noisy dataset variants.}

In addition, the WER score is computed for each transcript set $\widehat{T_{i_j}}$ with respect to references $T$. Accordingly, this produces WER scores $\{\{w_{i_j}\}_{j=0}^m\}_{i=0}^k$ (see Appendix \ref{sec_appendix_implementation_wer} for details). Notice that $T$'s WER is $0$. With the task scores and respective WER scores, we can now assess and compare the performance of the dataset variants.

\subsection{Analyzing the Results}
\label{sec_framework_analysis}

Each of the WER and task score-pairs $(w_{i_j}, s_{i_j})$ is a data point that can be plotted on a graph.
The curve $l_j = [(0, s)] \cdot [(w_{i_j}, s_{i_j})]_{i=0}^k$ describes the behavior of a task model as noise increases in the transcripts (as $i$ increases), when applying cleaning technique $j$ (or when no cleaning is enforced, at $j=0$).
These curves form a basis for analyzing the configured SLU pipeline, as explained next.
(See \autoref{fig_framework_graph} in the Appendix for visualization.)

\paragraph{Model performance vs. noise level.}
As transcript noise accumulates, NLU task model performance is expected to degrade.
One question to ask is: \textit{how much transcript noise can the task model tolerate before its performance is jeopardized?} To that end, we define the \textbf{\textit{noise-toleration point}} (NTP) as follows. For curve $l_j$, described by function\footnote{Note that the curve is not continuous since it is made up of several discrete segments. See Appendix \ref{sec_appendix_implementation_ntp} for details on how the noise-toleration point is computed.} $f_j$, and the respective upper and lower bound functions $f_j^{\text{upper}}$ and $f_j^{\text{lower}}$ (based on the margins-of-error), we define $l_j$'s noise-toleration point, $w^t_j$, as the WER score when $f_j^{\text{lower}}(0) = f_j^{\text{upper}}(w^t_j)$, i.e., 
the lowest WER at which the task score becomes statistically significantly lower than when transcripts have no noise, indicating a notable drop in task-model performance due to noise.

Another question to ask about the SLU pipeline is: \textit{how do different models behave comparatively, with respect to noise level?} The general behavior is approximated with the \textbf{\textit{area-under-the-curve}} (AUC), which can be compared between curves to judge which model is generally more tolerant to noise. Furthermore, by focusing on a certain region in the graph, the localized behavior is comparable. For example, in \autoref{fig_noclean_graphs}a, the GPT model is the better model at lower WER levels, but drops to the bottom rank at high WER levels.\footnote{The reliability of the analyses increases with the number of points constructing a curve (increasing $k$) and with a broader coverage of the WER score range (between 0 and 1).}


\paragraph{Comparing cleaning techniques.}
Applying a cleaning technique on transcripts decreases the noise, and consequently shifts the plots leftward. Cleaning a transcript also essentially means that the \textit{type} of noise changes, and therefore the task model reacts differently to the errors in the transcripts, potentially altering the behavior of the curves altogether.
The point $(w_{i_j}, s_{i_j})$ with respect to point $(w_{i_0}, s_{i_0})$ portrays how much ``effort'' is required (the decrease in WER: $w_{i_0} - w_{i_j}$) in order to change the task score from $s_{i_0}$ to $s_{i_j}$. The effect of each cleaning method $j$ varies, and therefore all $l_j$s are compared with respect to $l_0$ (e.g., see \autoref{fig_cleaning_graphs}). Ultimately, an effective cleaning technique should increase the task scores with minimum effort.

Formally, let $\Delta w_{i_j} = w_{i_0} - w_{i_j}$ be the change in WER for noising level $i$ and cleaning method $j$, and $\delta s_{i_j} = (s_{i_0} - s_{i_j}) / s$ be the respective relative\footnote{The change in task-score is normalized by the score at WER=0 to get the relative change. The change in WER is already on a 0-to-1 scale, and is not further normalized.} change in the task-score. The pointwise effectiveness score of cleaning technique $j$ at noise-level $i$ is measured as $e_{i_j} = \delta s_{i_j} / \sqrt{\Delta w_{i_j} + \epsilon}$.\footnote{We applied a square root transformation on the \textit{effort} ($\Delta w_{i_j}$) to reduce the impact of the larger changes at noisier levels, and to increase the weight of the change in task score ($\delta s_{i_j}$). $\epsilon$ is a minuscule value to prevent division by zero.} Finally, we measure the \textit{\textbf{cleaning-effectiveness score}} (CES) of cleaning method $j$ with the average: $\frac{1}{k+1} \sum_{i=0}^k e_{i_j}$. The higher the score, the better the overall improvement in the downstream task with a lower effort of cleaning. A score of 0 means that the cleaning procedure had no effect on the task-model's results, and a negative score means that there was a deterioration of task results, on average. 

The CES metric captures the two objectives of a cleaning technique: heightened task results for lesser effort.
The metric suggests how comparably effective a cleaning method is for the data and task-model in question. As such, it compares the effects of different \textit{types} of noise in the transcripts, as we exemplify in our experiments in Section \ref{sec_results}.


%% file: 04_setup.tex
\section{Experimental Setup}
\label{sec_setup}

\input{tables/datasets}

To demonstrate the utility of the framework for measuring \ENDow{}, we describe the various SLU pipeline configurations on which we apply the framework and conduct analyses (discussed in \S\ref{sec_results}).

\subsection{Preparing Transcript Sets}
\label{sec_setup_transcripts}

\paragraph{Text-to-speech model.}
Some of the SLU datasets in our experiments lack accompanying audio files, and in any case, we would like our experiments to be based on a controlled speech environment. We used the \texttt{toirtoise-tts} \citep{betker2023tortoisetts} Python library\footnote{\url{https://github.com/neonbjb/tortoise-tts}} as the text-to-speech model, and implemented a procedure for handling lengthy speech (see Appendix \ref{sec_appendix_implementation_tts}). The TTS stage produces the initial set of audio files for each of the SLU datasets in our experiments. 

\paragraph{Noising method.}
Each audio file was reverberated with the \texttt{rir-generator} \citep{werner2023rirgenerator} Python library,\footnote{\url{https://github.com/audiolabs/rir-generator}} and then recreated with background office sounds \citep[a clipped audio file;][]{myNoise2020office} with one of five signal-to-noise ratios (see Appendix \ref{sec_appendix_implementation_noising}). 
After this process there are six sets of increasingly tampered audio files.

\paragraph{ASR system.}
We used Whisper \citep{Radford2023whisper}\footnote{\texttt{openai/whisper-small.en}} for conducting speech-to-text (see Appendix \ref{sec_appendix_implementation_stt}). 
In all there are seven sets of increasingly noised transcripts (the first is the clean reference set). In our setting, seven noise levels provided a satisfactory analysis for examining the behavior of the SLU pipeline. The WER scores distribute within 0 and 0.9, and the curves empirically exhibit sufficiently clear behavioral patterns.

\paragraph{Cleaning techniques.}
In our experiments, we use the cleaning component to study the effect of different types of words, e.g., nouns, on downstream tasks. This analysis also simulates an SLU pipeline in which the ASR system prioritizes accuracy for specific word types, guiding where to focus efforts in the transcription process.

To clean transcripts, we first aligned a noised transcript to its respective reference transcript with \texttt{jiwer}.\footnote{\url{https://github.com/jitsi/jiwer}} Then, any non-equivalent alignment that involves the targeted word-type was repaired. We separately target nouns, verbs, adjectives, adverbs, any of the above (``content words''), none of the above (``non-content words''), and named entities -- seven techniques in all. Details in Appendix \ref{sec_appendix_implementation_cleaning}.

\subsection{Downstream Tasks}
\label{sec_setup_tasks}

We experiment with three downstream tasks, characterized by different output objectives. Summarization is a generation task where text is synthesized based on the collective understanding of a dialog. Question-answering is framed here as an extraction task that retrieves spans from the transcript. Dialog-act categorization is a classification task that assigns a communicative goal label (e.g., `statement', `question', etc.) to conversational utterances. The first two tasks are on the full transcript level, while the latter task is on the utterance level. These differences offer insights into potential distinctions in SLU pipelines.

In our experiments we focus on \textit{long spoken dialogues}, as opposed to short or written dialogues, as they impose a more challenging setting for task models.
See \autoref{tab_datasets} for a summary of the tasks, and Appendix \ref{sec_appendix_datasets} for examples of task instances.

\paragraph{Summarization.}
For summarization, we use the QMSum dataset\footnote{\url{https://github.com/Yale-LILY/QMSum}} \citep{zhong-etal-2021-qmsum}, a generic and query-focused dialog summarization benchmark. It consists of transcripts and summaries of product meetings \citep[AMI;][]{carletta2006ami}, academic meetings \citep[ICSI;][]{janin2003icsi} and parliament committee meetings.

To evaluate system summaries we use standard ROUGE metrics\footnote{\url{huggingface.co/spaces/evaluate-metric/rouge}, with the default arguments.} \citep{lin-2004-rouge} and pairwise comparison ranking \citep{qin-etal-2024-large} with \texttt{GPT-4o-mini} as a judge for overall quality \citep{liu-etal-2024-benchmarking} (see Appendix \ref{sec_appendix_implementation_pairwise} for details).

\paragraph{Question-answering.}
The QAConv dataset\footnote{\url{https://github.com/salesforce/QAConv}} \citep{wu-etal-2022-qaconv} consists of dialogues with questions whose answers are short spans in the dialog. We only use the instances based on court cases or interviews (since these are long spoken dialogues).

For evaluation, predicted answers are compared against reference answers with exact match accuracy, token-level $F_1$ and fuzzy matching, following the QAConv benchmark.

\paragraph{Dialog-act classification.}
The MRDA dataset\footnote{\url{https://github.com/NathanDuran/MRDA-Corpus}} \citep{shriberg-etal-2004-icsi} consists of meetings from the ICSI corpus and research-oriented group meetings. Each utterance in the transcripts is labeled with one of 12 dialog act labels \citep{dhillon2004mrdaLabeling}. We utilize the first and last 50 utterances from each transcript (100 of $\sim$1392), for efficiency purposes. See Appendix \ref{sec_appendix_datasets} for more details.

The MRDA results were traditionally evaluated with the accuracy metric, but we also evaluate with macro-$F_1$ due to the high class imbalance in the dataset, as suggested by \citet{miah-etal-2023-hierarchical}.

\paragraph{Models.}
For all three tasks, we experiment with four instruct-tuned LLMs in zero-shot mode:
Mistral-7B,\footnote{\texttt{mistralai/Mistral-7B-Instruct-v0.3}} Llama3-8B,\footnote{\texttt{meta-llama/Meta-Llama-3-8B-Instruct}} Llama3.1-8B,\footnote{\texttt{meta-llama/Llama-3.1-8B-Instruct}} and GPT-4o-mini.\footnote{\texttt{gpt-4o-mini-2024-07-18}} 
They were selected for their modest hardware requirements and affordability.
Since the context size of Mistral and Llama-3 cannot fit most of the transcripts in full, summarization and QA were conducted on these models in segments.
See details and prompts in Appendix \ref{sec_appendix_prompts}.

%% file: tables/datasets.tex
\begin{table*}[t]
\centering
\resizebox{\textwidth}{!}{%
\begin{tabular}{@{}lcccccccc@{}}
\toprule
\textbf{Task} & \makecell[c]{\textbf{Dataset} \\ \textbf{Source}} & \textbf{Task Type} & \makecell[c]{\textbf{Granularity} \\ \textbf{Level}} & \textbf{Domains} & \makecell[c]{\textbf{Evaluation} \\ \textbf{Metrics}} & \makecell[c]{\textbf{\#} \\ \textbf{Transcripts}} & \makecell[c]{\textbf{\#} \\ \textbf{Instances}} & \makecell[c]{\textbf{\# Utts. in Inst.} \\ \textbf{Avg. (min - max)}} \\
\midrule
\makecell[l]{Summarization\\(Generic + Query Focused)} & QMSum & Generation & Transcript & \makecell[c]{Product, Research, \\Parliament meetings} & \makecell[c]{Pairwise Ranking, \\ROUGE-1,2,L} & \makecell[c]{35} & \makecell[c]{281} & \makecell[c]{592 \\ (131 - 1368)} \\
\midrule
Question Answering & QAConv & Extraction & Transcript & \makecell[c]{Court cases, \\ Interviews} & \makecell[c]{Fuzzy match, Exact,\\ Token-level $F_1$} & \makecell[c]{505} & \makecell[c]{2083} & \makecell[c]{104 \\ (5 - 585)} \\
\midrule
Dialog Act Classification & MRDA & Classification & Utterance & Research meetings & Macro-$F_1$, Accuracy & \makecell[c]{12} & \makecell[c]{1200} & \makecell[c]{1} \\
\bottomrule
\end{tabular}%
}
\caption{The test data with which we conduct our experiments, with varying input/output formats and speech domains.
\textit{\# Instances} refers to queries for summarization, questions for QA, and utterances for dialog-act classification.}
\label{tab_datasets}
\end{table*}

%% file: 05_results.tex
\section{Results and Analyses}
\label{sec_results}

\input{figures/noclean/noclean_main}

Our experiments include the configurations described in Section \ref{sec_setup}, with results discussed here.

\paragraph{Comparing task models.}
The following analyses reveal how task models perform under varying levels of noise. The AUC and NTP scores provide a high-level comparison for assessing \ENDow{}, while the graph curves illustrate model behavior as noise fluctuates.

\autoref{fig_noclean_graphs}
presents the results on the three downstream tasks for the four models, based on one of the evaluation metrics per task.\footnote{Graphs based on the other metrics are in \autoref{fig_noclean_graphs_all} in the Appendix. The analysis here is for demonstration purposes; additional insights could be gathered from the other graphs.} 
In the summarization task results (\autoref{fig_noclean_graphs}a) we first notice that models tend to tolerate a noise level of about 0.2 WER (NTP between 0.07 and 0.3), i.e., task scores are not significantly lower ($p<0.05$) until that level of noise. Also, while the AUC scores for all models are not significantly different ($p<0.05$), models behave differently with respect to WER. For example, GPT's summaries are more highly preferred at lower WER than at higher WER values, while for Mistral, the preference is slightly more evenly distributed.

In the question-answering task (\autoref{fig_noclean_graphs}b), the more advanced models (GPT and Llama-3.1) yield substantially better results than the other two models. This could be an effect of the small context window which requires conducting the task in segments, likely inducing more errors. 
Similar to the behavior in summarization, here too GPT yields better scores than Llama 3.1 at low WER, but Llama 3.1 surpasses GPT as WER increases.

In the dialog-act classification task (\autoref{fig_noclean_graphs}c), the noise-toleration points are quite high due to the large margins-of-error and relatively flat curves. A high NTP either implies that noise has little effect on a downstream task, or alternatively that the model is ineffective for the task in general.
In this case, the latter seems to be the case, when comparing to results of \citet{miah-etal-2023-hierarchical} ($0.29$ macro-$F_1$ and $0.6$ accuracy vs. $[0.17, 0.32]$ macro-$F_1$ and $[0.3, 0.48]$ accuracy here). More in Appendix \ref{sec_appendix_results}.


\input{figures/cleaning/cleaning_main}

\input{tables/scores_cleaning}

\paragraph{Comparing noise types.}
The following analyses highlight the impact of different \textit{types} and intensities of noise. This examination enables more efficient optimization of ASR systems and post-hoc transcript repair. For instance, as the analysis will show, prioritizing named-entity accuracy may be more effective than generally minimizing the WER of the ASR system.

\autoref{fig_cleaning_graphs} presents the graphs showing the effect of cleaning techniques on the performance of GPT (graphs for the rest of the models and cleaning techniques in Figures \ref{fig_cleaning_graphs_all_qmsum}, \ref{fig_cleaning_graphs_all_qaconv} and \ref{fig_cleaning_graphs_all_mrda} in the Appendix). The ``no cleaning'' curve shows model results on the transcripts at the various noise levels. The other curves on the graph show model results when also applying a cleaning technique on the same transcripts.

In the summarization task (\autoref{fig_cleaning_graphs}a), fixing all the content words in the transcripts (``content'' curve) helps the model produce summaries that are preferred over all the summaries that are based on the original transcripts, regardless of noise level. However, due to this technique's costly ``effort'' (high change in WER), the cleaning-effectiveness score (0.479) is not as high as that of the ``named entities'' technique (0.499). The latter cleaning method improves the task scores at a smaller cost of effort on average, as depicted in the graph. These findings indicate the value of content words, and named entities in particular, for the summarization task using GPT. Notice that transcripts that are almost fully error-prone (WER is $\sim$0.9) are fixed to a WER of $\sim$0.4 by repairing content words, but resulting summaries are much preferred over summaries from transcripts with different types of errors, also at a WER of 0.4. This further stresses the importance of analyzing the \textit{types} of errors in transcripts and not just the \textit{amount}, which is a known limitation of the WER metric.

\autoref{tab_scores_cleaning} lists the CES scores of each model and cleaning technique, ranked by the effectiveness on GPT.
For the QA task, we see that the cleaning techniques are ranked quite similarly across the four task-models. In all three tasks, fixing only adverbs or verbs is less effective. On the other end, nouns, and named-entities particularly, are more effective for the transcript-level tasks (summarization and QA). The utterance-level task of dialog-act classification behaves differently with respect to the repaired word-types. Interestingly, repairing non-content words is effective for the task, consistent with works that found that function words are essential features for classifying dialog-acts \citep{OShea2012daFunction, jo-etal-2017-modeling}.
Named-entities are also effective for the task with GPT. A closer look into the graph (\autoref{fig_cleaning_graphs}c) reveals that the high CES is affected by the behavior at lower WER scores, where a strong increase in the task score is obtained at a small effort.

%% file: figures/noclean/noclean_main.tex
\begin{figure}[h!]
    \centering
    \subfloat[QMSum with pairwise ranking evaluation]{
        \includegraphics[width=0.9\columnwidth, trim=8 8 5 20, clip]{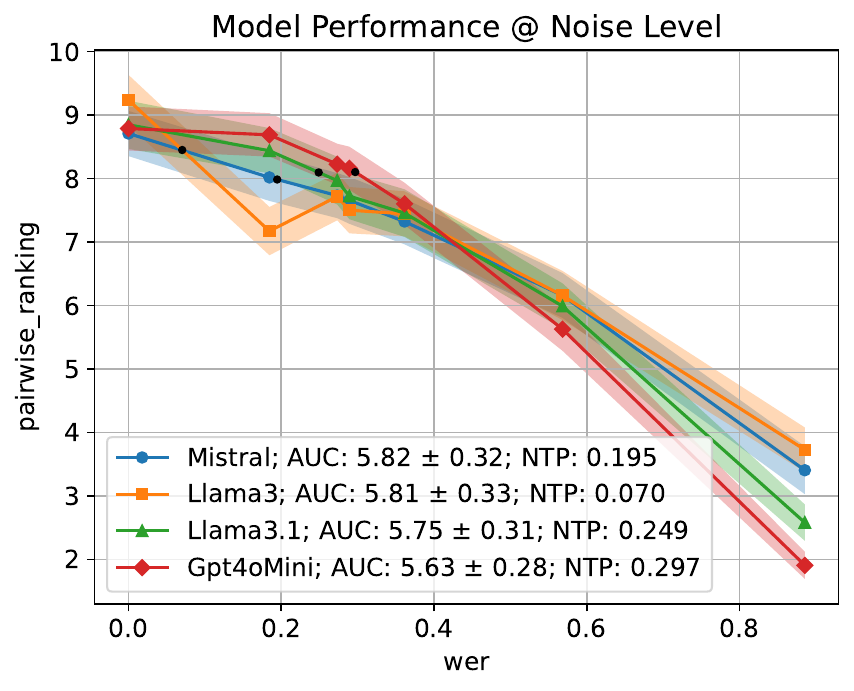}
    }
    \hspace{0.2cm}
    \subfloat[QAConv with fuzzy match evaluation]{
        \includegraphics[width=0.9\columnwidth, trim=8 8 5 20, clip]{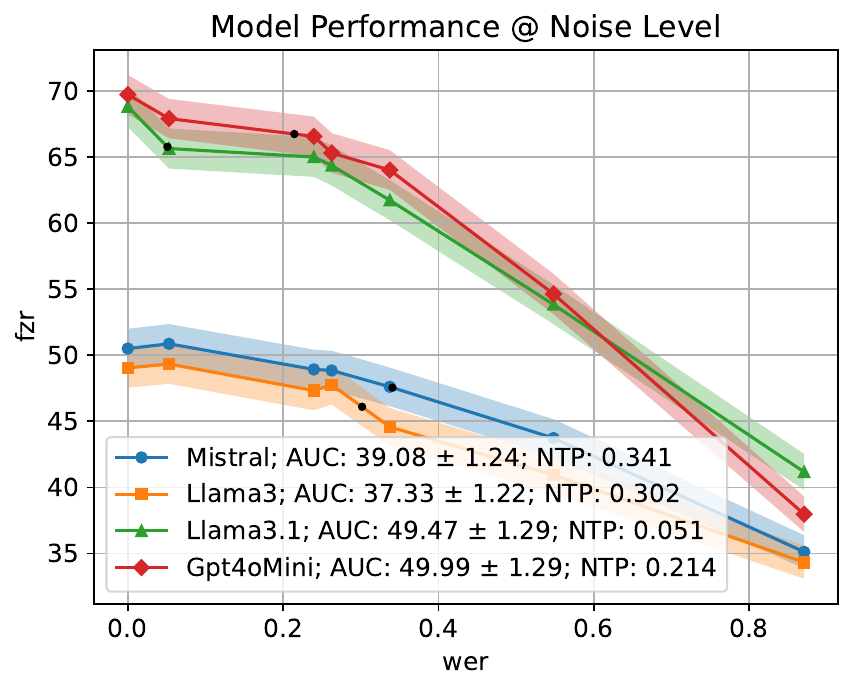}
    }
    \hspace{0.2cm}
    \subfloat[MRDA with macro-$F_1$ evaluation]{
        \includegraphics[width=0.9\columnwidth, trim=8 8 5 20, clip]{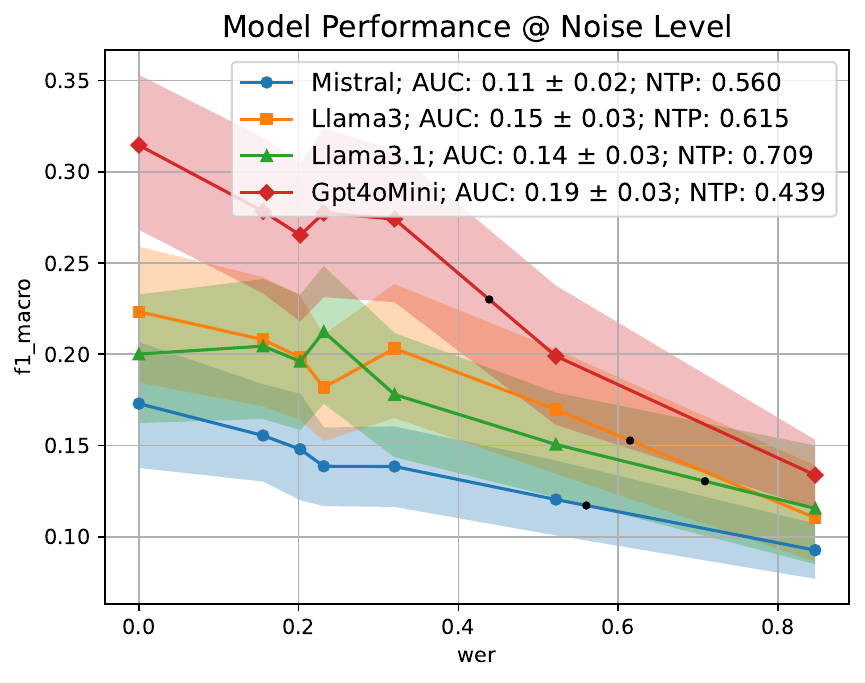}
    }
    \caption{Model performance on the experimented tasks.
    Curves are compared with area-under-the-curve (AUC) and noise-toleration points (NTP; marked with black dots). NTP marks the WER value where the task-score first decreases significantly from the score at $\text{WER}=0$. A line's shaded area represents its confidence interval. Graphs for the rest of the metrics are in \autoref{fig_noclean_graphs_all}.}
    \label{fig_noclean_graphs}
\end{figure}

%% file: figures/cleaning/cleaning_main.tex
\begin{figure}[h!]
    \centering
    \subfloat[QMSum with pairwise ranking evaluation, for GPT]{
        \includegraphics[width=0.9\columnwidth, trim=8 8 14 21, clip]{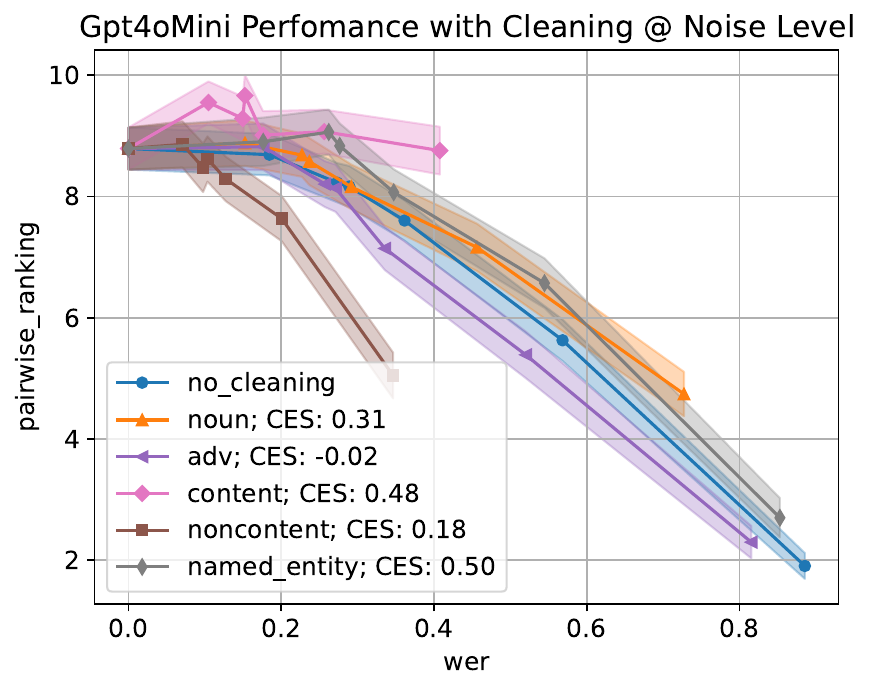}
    }
    \hspace{0.2cm}
    \subfloat[QAConv with fuzzy matching evaluation, for GPT]{
        \includegraphics[width=0.9\columnwidth, trim=8 8 14 21, clip]{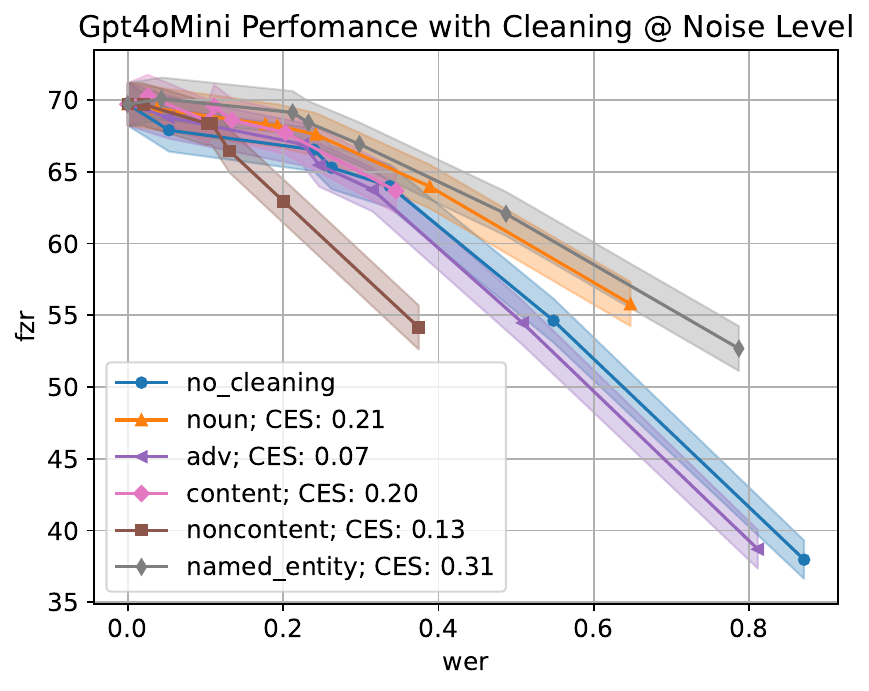}
    }
    \hspace{0.2cm}
    \subfloat[MRDA with macro-$F_1$ evaluation, for GPT]{
        \includegraphics[width=0.9\columnwidth, trim=8 8 14 21, clip]{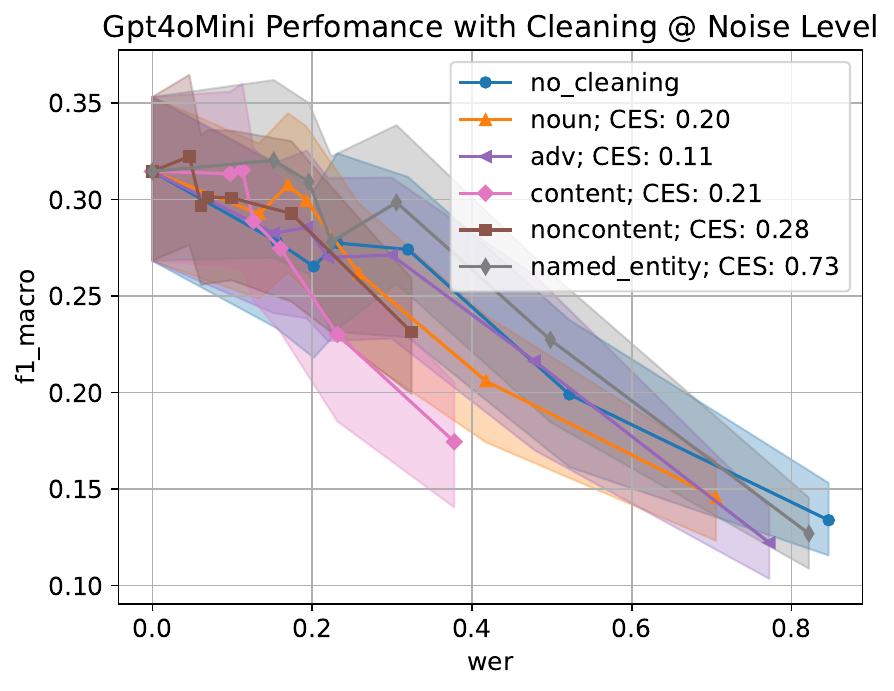}
    }
    \caption{The performance of GPT-4o-mini when applying various cleaning techniques. Compare a point on the ``no\_cleaning'' curve to the respective point on a cleaning technique's curve. Effective cleaning means maximizing gain in task score (y-axis) with minimum effort (x-axis), measured using the cleaning-effectiveness score (CES).
    Additional CES scores are in \autoref{tab_scores_cleaning}, and more graphs are in Figures \ref{fig_cleaning_graphs_all_qmsum}, \ref{fig_cleaning_graphs_all_qaconv} and \ref{fig_cleaning_graphs_all_mrda} in the Appendix.}
    \label{fig_cleaning_graphs}
\end{figure}

%% file: tables/scores_cleaning.tex
\begin{table}[t]
    \centering
    \resizebox{\columnwidth}{!}{%
    \begin{tabular}{c|lc|c|c|c}
    \toprule
 & & Mistral & Llama3 & Llama3.1 & GPT4oMini \\
    \midrule
    \multirow{7}{*}{\rotatebox[origin=c]{90}{QMSum | PW-Rank}}
    &
    Named-ents &     \cellcolor[gray]{1.00} \textcolor{black}{0.537} &     \cellcolor[gray]{1.00} \textcolor{black}{1.346} &     \cellcolor[gray]{0.33} \textcolor{white}{0.193} &     \cellcolor[gray]{1.00} \textcolor{black}{0.499} \\
    &
    Content &     \cellcolor[gray]{0.50} \textcolor{white}{0.322} &     \cellcolor[gray]{0.17} \textcolor{white}{0.581} &     \cellcolor[gray]{0.83} \textcolor{black}{0.357} &     \cellcolor[gray]{0.83} \textcolor{black}{0.479} \\
    &
    Nouns &     \cellcolor[gray]{0.83} \textcolor{black}{0.459} &     \cellcolor[gray]{0.50} \textcolor{white}{0.804} &     \cellcolor[gray]{1.00} \textcolor{black}{0.384} &     \cellcolor[gray]{0.67} \textcolor{black}{0.305} \\
    &
    Non-content &     \cellcolor[gray]{0.33} \textcolor{white}{0.262} &     \cellcolor[gray]{0.00} \textcolor{white}{0.480} &     \cellcolor[gray]{0.50} \textcolor{white}{0.209} &     \cellcolor[gray]{0.50} \textcolor{white}{0.181} \\
    &
    Adjectives &     \cellcolor[gray]{0.67} \textcolor{black}{0.327} &     \cellcolor[gray]{0.83} \textcolor{black}{1.216} &     \cellcolor[gray]{0.67} \textcolor{black}{0.230} &     \cellcolor[gray]{0.33} \textcolor{white}{0.135} \\
    &
    Verbs &     \cellcolor[gray]{0.17} \textcolor{white}{0.229} &     \cellcolor[gray]{0.33} \textcolor{white}{0.702} &     \cellcolor[gray]{0.17} \textcolor{white}{0.145} &     \cellcolor[gray]{0.17} \textcolor{white}{0.073} \\
    &
    Adverbs &     \cellcolor[gray]{0.00} \textcolor{white}{0.223} &     \cellcolor[gray]{0.67} \textcolor{black}{0.956} &     \cellcolor[gray]{0.00} \textcolor{white}{0.017} &     \cellcolor[gray]{0.00} \textcolor{white}{-0.023} \\
    \midrule
    \multirow{7}{*}{\rotatebox[origin=c]{90}{QAConv | Fuzzy}}
    &
    Named-ents &     \cellcolor[gray]{1.00} \textcolor{black}{0.221} &     \cellcolor[gray]{1.00} \textcolor{black}{0.469} &     \cellcolor[gray]{1.00} \textcolor{black}{0.294} &     \cellcolor[gray]{1.00} \textcolor{black}{0.311} \\
    &
    Nouns &     \cellcolor[gray]{0.83} \textcolor{black}{0.164} &     \cellcolor[gray]{0.50} \textcolor{white}{0.280} &     \cellcolor[gray]{0.83} \textcolor{black}{0.210} &     \cellcolor[gray]{0.83} \textcolor{black}{0.211} \\
    &
    Content &     \cellcolor[gray]{0.67} \textcolor{black}{0.108} &     \cellcolor[gray]{0.33} \textcolor{white}{0.224} &     \cellcolor[gray]{0.67} \textcolor{black}{0.186} &     \cellcolor[gray]{0.67} \textcolor{black}{0.202} \\
    &
    Non-content &     \cellcolor[gray]{0.50} \textcolor{white}{0.070} &     \cellcolor[gray]{0.00} \textcolor{white}{0.177} &     \cellcolor[gray]{0.33} \textcolor{white}{0.094} &     \cellcolor[gray]{0.50} \textcolor{white}{0.133} \\
    &
    Adjectives &     \cellcolor[gray]{0.00} \textcolor{white}{-0.037} &     \cellcolor[gray]{0.83} \textcolor{black}{0.398} &     \cellcolor[gray]{0.50} \textcolor{white}{0.109} &     \cellcolor[gray]{0.33} \textcolor{white}{0.120} \\
    &
    Verbs &     \cellcolor[gray]{0.17} \textcolor{white}{-0.012} &     \cellcolor[gray]{0.17} \textcolor{white}{0.188} &     \cellcolor[gray]{0.17} \textcolor{white}{0.059} &     \cellcolor[gray]{0.17} \textcolor{white}{0.090} \\
    &
    Adverbs &     \cellcolor[gray]{0.33} \textcolor{white}{0.004} &     \cellcolor[gray]{0.67} \textcolor{black}{0.335} &     \cellcolor[gray]{0.00} \textcolor{white}{0.033} &     \cellcolor[gray]{0.00} \textcolor{white}{0.071} \\
    \midrule
    \multirow{7}{*}{\rotatebox[origin=c]{90}{MRDA | Mac-$F_1$}}
    &
    Named-ents &     \cellcolor[gray]{0.17} \textcolor{white}{-0.035} &     \cellcolor[gray]{0.17} \textcolor{white}{-0.049} &     \cellcolor[gray]{0.17} \textcolor{white}{-0.291} &     \cellcolor[gray]{1.00} \textcolor{black}{0.735} \\
    &
    Adjectives &     \cellcolor[gray]{1.00} \textcolor{black}{0.404} &     \cellcolor[gray]{0.50} \textcolor{white}{-0.001} &     \cellcolor[gray]{0.83} \textcolor{black}{0.035} &     \cellcolor[gray]{0.83} \textcolor{black}{0.290} \\
    &
    Non-content &     \cellcolor[gray]{0.83} \textcolor{black}{0.392} &     \cellcolor[gray]{1.00} \textcolor{black}{0.122} &     \cellcolor[gray]{1.00} \textcolor{black}{0.122} &     \cellcolor[gray]{0.67} \textcolor{black}{0.285} \\
    &
    Content &     \cellcolor[gray]{0.67} \textcolor{black}{0.315} &     \cellcolor[gray]{0.67} \textcolor{black}{0.027} &     \cellcolor[gray]{0.50} \textcolor{white}{-0.053} &     \cellcolor[gray]{0.50} \textcolor{white}{0.212} \\
    &
    Nouns &     \cellcolor[gray]{0.33} \textcolor{white}{0.010} &     \cellcolor[gray]{0.83} \textcolor{black}{0.102} &     \cellcolor[gray]{0.67} \textcolor{black}{-0.042} &     \cellcolor[gray]{0.33} \textcolor{white}{0.203} \\
    &
    Verbs &     \cellcolor[gray]{0.50} \textcolor{white}{0.044} &     \cellcolor[gray]{0.33} \textcolor{white}{-0.024} &     \cellcolor[gray]{0.33} \textcolor{white}{-0.232} &     \cellcolor[gray]{0.17} \textcolor{white}{0.158} \\
    &
    Adverbs &     \cellcolor[gray]{0.00} \textcolor{white}{-0.043} &     \cellcolor[gray]{0.00} \textcolor{white}{-0.085} &     \cellcolor[gray]{0.00} \textcolor{white}{-0.342} &     \cellcolor[gray]{0.00} \textcolor{white}{0.107} \\
    \bottomrule
    \end{tabular}%
    }
    \caption{The cleaning-effectiveness scores (CES) of the experimented cleaning techniques on the four task-models.
    Techniques ordered for each task by ranking on GPT model (corresponding graphs in \autoref{fig_cleaning_graphs}). Full table for all task-metrics in \autoref{tab_scores_cleaning_all} in the Appendix.}
    \label{tab_scores_cleaning}
\end{table}

%% file: 06_conclusion.tex
\section{Conclusion}
\label{sec_conclusion}

Errors in speech-to-text automation propagate to downstream language understanding tasks, with noise magnitude and type affecting tasks and models differently. We present a configurable framework for evaluating noise impact, enabling analysis of model behavior across noise levels and transcript-cleaning techniques. Extensive experiments demonstrate its utility, providing insights into task model performance in SLU. The framework’s flexibility supports more effective comparison and development of SLU pipelines.

%% file: 07_limitations.tex
\section*{Limitations}

In our experiments, the pipelines are initiated with relatively clean and clear audio files, and the subsequent acoustic deterioration is done in a specific manner (reverberation and background sounds). Other acoustic settings are indeed possible for initiating the SLU pipeline, e.g., with a low-resourced lingual dialect, different speaker voices per turn, overlapping speech, microphone settings, and many other parameters. Our framework is robust to these variants, and the purpose of our experiments is to exemplify the utility of the framework.

Similarly, our experiments are limited to the configurations we defined, for demonstrating the framework. Other configurations could involve non-English languages, different tasks, models and SLU/NLU datasets. The resulting analyses could yield findings that are different from ours, which reiterates the need for a robust framework like ours.

The cleaning techniques we used depend on the reference transcript in order to identify the word/phrase types that we want to include in our analysis. Our experiments show how \textit{different types} of errors affect a downstream task. A cleaning technique can also be one that is used in practice without dependence on the reference transcript. In the latter case, our framework would indicate the effectiveness of a transcript-cleaning component within an SLU pipeline.



We emphasize that the behavior of a graph depends on the task metric applied, and the resulting analysis can therefore differ when using different metrics for the same task and data. When insights are gathered with the framework, it is important to strongly consider the metric used, or use several metrics to paint a fuller picture.

%% file: 0a_appendix.tex
\appendix

\section{Implementation Details}
\label{sec_appendix_implementation}

\subsection{Computing Overall WER on a Set of Transcripts}
\label{sec_appendix_implementation_wer}

Before computing WER, the utterances in a transcript are tokenized using \texttt{spaCy} \citep{spacy2017}, and then the tokens are recombined with separating spaces. This is done mainly to separate punctuation and contractions.

The WER score for a transcript is computed against the reference transcript using the \texttt{jiwer.process\_words} function. It computes hits, insertions, substitutions and deletions between each predicted and respective reference utterance. Then all of the utterance level values are added up, and a single WER score is computed accordingly for the transcript.

Once the transcript-level WER scores are computed, the transcript-set WER score is the average over all transcripts in the set.

\subsection{Computing the Noise-toleration Point (NTP)}
\label{sec_appendix_implementation_ntp}

Observe \autoref{fig_framework_graph} for a visualization of the following explanation.
The computation of the noise-toleration point of a curve $l_j$ relies on $l_j^{\text{upper}}$ and $l_j^{\text{lower}}$, the lines with the respective \textbf{margins-of-error}. The y-value, i.e., task score, of each point in $l_j^{\text{upper}}$ ($l_j^{\text{lower}}$) is the upper (lower) limit margin-of-error for the corresponding point's y-value in $l_j$. Specifically, task score $s_{i_j}$, as part of point $(w_{i_j}, s_{i_j})$ on $l_j$, is the average task score over the transcripts in the set $\widehat{T_{i_j}}$, and the respective margin-of-error is computed at a confidence level of $95\%$ with the formula $1.96*\sigma/\sqrt{n}$, where $\sigma$ is the standard deviation of the scores, and $n$ is the number of scores (number of transcripts). The x-values on $l_j^{\text{upper}}$ and $l_j^{\text{lower}}$ are kept the same as in the respective points on $l_j$. With these margins-of-error line, we can compute the NTP.

Since a model's curve $l_j$ is constructed of several discrete points, we find the first point $p_i:=(w_{i_j}, s_{i_j})$ in $l_j$ whose upper bound (based on the margin-of-error) is above the lower bound of the first point $p_0$, and where the next point $p_{i+1}$ has an upper bound that is below $p_0$'s lower bound. Then on the linear segment between $p_i$ and $p_{i+1}$, we find the $x$ value (WER score) where its upper bound is equal to the lower bound of $p_0$.

\input{figures/graph_illustration}

\subsection{Executing Text-to-speech}
\label{sec_appendix_implementation_tts}

To create the initial audio files from the dataset transcripts, we first removed substrings in the utterances that are within curly and box brackets. These are used in some transcripts to indicate non-verbal markers. We also removed redundant whitespaces. An utterance was sentence-tokenized (with \texttt{nltk.tokenize.sent\_tokenize}), and then also broken up to segments of up to 50 tokens, if it was longer than that (with \texttt{nltk.tokenize.word\_tokenize}). We found that in some cases the TTS module had some difficulty in voicing more than 50 tokens at a time (more than about 15 seconds of speech). Each segment was then passed to \texttt{tortoise.api.TextToSpeech} and an audio file was created with the ``emma'' voice in ``ultra\_fast'' mode, and saved with a 24000 sample frequency.

\subsection{Impairing Audio Files}
\label{sec_appendix_implementation_noising}

An audio file is recreated $k$ times at increasing levels of speech deterioration. First the audio file is reverberated once using the Room Impulse Response Generator \citep[\texttt{rir\_generator} library;][]{werner2023rirgenerator}. Then background sounds are added at $k$ different signal-to-noise ratios, as described below.

The reverberation parameters are as follows. The \textbf{room dimensions} are uniformly selected for each of width (2 to 10 meters), length (2 to 10 meters) and height (exactly 3 meters). Assuming that the room is enclosed by walls, a floor and a ceiling, the \textbf{speaker position} is uniformly selected for each of x-position (somewhere 0.5 meters from the wall), y-position (somewhere 0.5 meters from the wall), and z-position (somewhere between the floor and the ceiling). The \textbf{microphone location} is randomly placed 2 meters away from the speaker if it's within the bounds of the room, otherwise 1 meter away, otherwise 0 meters away. The \textbf{Reverberation Time} (RT60 -- the time it takes for sound energy to decrease by 60 dB after the sound source stops) is uniformly selected between 0.15 and 1 second. \textbf{Sound velocity} is kept at the default value of 340 meters per second. The \textbf{sample frequency} is kept at the original value (24000 samples per second).

To the resulting reverberated audio file denoted \texttt{signal}, \textbf{background sounds} are added at $k$ different signal-to-noise ratios (SNR -- level of a desired signal to the level of background noise). First a sound audio file \citep{myNoise2020office} denoted \texttt{noise\_signal} (in our case we used an office background that includes realistic sounds such as chatter, papers, office machinery, drinking, etc.) is loaded, and repeated so that its length is equal to that of the speech audio file, or truncated to that length. The resulting background file is denoted \texttt{white\_noise}. Then the noise factor is computed according to the SNR with:
\[
g = \sqrt{\frac{10^{-\texttt{SNR}/10} * \text{std}(\texttt{signal})^2}{\epsilon + \text{std}(\texttt{noise\_signal})^2}}
\]
The final audio file is created as:
\[
\texttt{noisy\_signal} = \texttt{signal} + g * \texttt{white\_noise}
\]
The $k$ SNR values that we use in our experiments are -10, -5, 0, 5, 10. The higher the value, the more distinctive the speech is over the background noise.

\subsection{Executing ASR for Speech-to-text}
\label{sec_appendix_implementation_stt}

To run Whisper on a transcript, a Huggingface \citep{wolf2020huggingfaces} automatic-speech-recognition pipeline is initialized with the \texttt{openai/whisper-small.en} model. The pipeline receives each audio file and generates the respective text. Since our audio files are up to about 15 seconds in length and mostly under 1MB in size, the model is able to handle the files properly.

\subsection{Cleaning Transcripts}
\label{sec_appendix_implementation_cleaning}

Our cleaning techniques rely on the reference transcripts, and therefore are used to inidicate how different \textit{types} of errors effect the behavior of a model on a downstream task.

Given a chunk of 20 utterances from the predicted transcript, and the respective chunk from the reference transcript, \texttt{spaCy} is used to tag the part-of-speech labels of each token, and the named entity chunks. Then \texttt{jiwer.process\_words} is used to align the texts. For each alignment, if there is a substitution, addition or deletion, and it involves a type (POS or named entity) that is being cleaned, then that alignment is fixed.

As mentioned, we separately clean each of: nouns, verbs, adjectives, adverbs, all the above (content words), none of the above (non-content words) and named entities.

An example for cleaning nouns (in bold):

\noindent
Reference: \textit{``We certainly see it, as employers. The penny drops after a few weeks or months.''}

\noindent
Noisy: \textit{``We certainly seen it as \textbf{lawyers}. The penny drops after the new \textbf{songs}.''}

\noindent
Cleaned: \textit{``We certainly seen it as \textbf{employers} . The penny drops after the new \textbf{weeks months}.''}

\subsection{Pairwise Ranking for Summarization Evaluation}
\label{sec_appendix_implementation_pairwise}

Pairwise comparison has been shown to be highly effective for judging the overall quality of summaries \citep{liu-etal-2024-benchmarking} using \texttt{gpt-4-0314} as a judge. We use a presumably more advanced model (\texttt{gpt-4o-mini}), and assume high reliability.

Given two competing summaries, the reference summary, and an optional query on which the summary is focused, a pairwise comparer needs to mark the preferred summary (in our case, for ``general quality''). The two summaries being compared are presented to the comparer in random order to remove an order bias with regard to the preference made. We use a method inspired by the LLMCompare protocol from \citet{liu-etal-2024-benchmarking}.

For a generic summary we input the following prompt to \texttt{gpt-4o-mini}:

\begin{tcolorbox}[colback=gray!10, colframe=gray!50, sharp corners, boxrule=0.5mm]
\small{
\texttt{You will be given a generic reference summary of a conversation, as well as two summaries written by automatic systems. Your task is to decide which of the two system summaries is better, with respect to the reference summary. If it is difficult to decide which summary has better overall quality, then you may say that there is a tie. \\
First explain briefly the reasoning for your choice, and then provide an answer as 1, 2 or tie. \\
The output should be in the following format:\\
Explanation: <your reasoning>\\
Response: <1, 2 or tie>\\\\
Reference summary: \{summ\_ref\}\\
System 1 summary: \{summ\_1\}\\
System 2 summary: \{summ\_2\}}}
\end{tcolorbox}

For a query-focused summary we input the following prompt to \texttt{gpt-4o-mini}:

\begin{tcolorbox}[colback=gray!10, colframe=gray!50, sharp corners, boxrule=0.5mm]
\texttt{You will be given a query-focused reference summary of a conversation, as well as two summaries written by automatic systems. Your task is to decide which of the two system summaries is better, with respect to the reference summary and the query. If it is difficult to decide which summary has better overall quality, then you may say that there is a tie. \\
First explain briefly the reasoning for your choice, and then provide an answer as 1, 2 or tie. \\
The output should be in the following format:\\
Explanation: <your reasoning>\\
Response: <1, 2 or tie>\\\\
Query: \{query\} \\
Reference summary: \{summ\_ref\}\\
System 1 summary: \{summ\_1\}\\
System 2 summary: \{summ\_2\}}
\end{tcolorbox}

\paragraph{Pairwise ranking for non-cleaned versions.}
A non-cleaned transcript-set is prepared at seven levels of noise (no noise, and six levels of increasing intensity). These are marked as $T$ (the reference transcript set) and $\{\widehat{T_{i_0}}\}_{i=0}^k$ (noisy versions) respectively. The resulting summary sets are denoted $\widehat{O}$ (on the reference transcripts $T$) and $\{\widehat{O_{i_0}}\}_{i=0}^k$. Then the summaries for instance $z$, i.e., $\widehat{o_{i_0}^z} \in \widehat{O_{i_0}}$ and $\widehat{o^z} \in \widehat{O}$ (seven summaries total) are compared in pairs through the pairwise compairson LLM annotation. In all there are 21 pairs, and the preferred summary receives 2 points, or 1 point for a tie. Hence, a summary that is preferred over all other six summaries can score a maximum of 12 points, and a total of 42 points are distributed amongst the seven summaries. The scores for each noise version are averaged over all instances of the test set (281 instances), and margin-of-errors are computed. These scores are then plotted on the curve, e.g., as in \autoref{fig_noclean_graphs}a.

Using the Azure OpenAI API,\footnote{\url{https://azure.microsoft.com/en-us/pricing/details/cognitive-services/openai-service/}} the cost for computing all comparisons for one model was about \$1.50. For the four models this process totalled about \$6.

\paragraph{Pairwise ranking for cleaning techniques.}
For a cleaning technique $j$, a summary $\widehat{o_{i_j}^z} \in \widehat{O_{i_j}}$ is instead compared to summaries $\widehat{o_{i_0}^z} \in \widehat{O_{i_0}}$ and $\widehat{o^z} \in \widehat{O}$. In this case, we want to assess how the cleaned summary compares against the non-cleaned summaries. Each of the six cleaned summaries is compared against seven non-cleaned summaries, for a total of 42 comparisons per instance. A summary can score up to 14 points, and 84 points are distributed amongst the compared summaries. The scores for the cleaned summaries are avarged over all instances of the test set (281 instances), and margin-of-errors are computed. These scores are then plotted on the curve, e.g., as in \autoref{fig_cleaning_graphs}a. In this graph, the non-cleaned line is reused from the non-cleaned pairwise ranking from before, except that it is shifted up one point in the y-axis. This is done as if to emulate the same procedure done here where the non-cleaned summaries should be compared to \textit{all} the non-cleaned summaries (including itself), and would hence receive an additional point for the tie of a summary against itself.

Using the Azure OpenAI API, the cost for computing all comparisons for one model and one cleaning method was about \$4.30. For the four models and seven cleaning techniques, this amounted to about \$120.





\section{Executing Task Models}
\label{sec_appendix_prompts}

The four LLMs in our experiments were executed in zero-shot mode. \texttt{mistral-7b}, \texttt{llama-3-8b} and \texttt{llama-3.1-8b} were run on a local server. \texttt{gpt-4o-mini} was run with the Azure OpenAI API.

The summarization and question-answering tasks are given a transcript in the input. The transcripts and therefore chunked to fit (with the instruction) the context window of the employed model. For \texttt{mistral-7b} and \texttt{llama-3-8b} the size is 8K, for \texttt{llama-3.1-8b} and \texttt{gpt-4o-mini} it is 128K. The latter two models do not require chunking for our tested data. The prompts were scripted with some light prompt engineering on a few instances.

\subsection{Summarization with QMSum}
\label{sec_appendix_prompts_qmsum}

The prompt for summarizing the full transcript is:

\begin{tcolorbox}[colback=gray!10, colframe=gray!50, sharp corners, boxrule=0.5mm]
\texttt{Given the following conversation, answer the question: \{query\}\\
The conversation is:\\
\{transcript\}}
\end{tcolorbox}

\noindent
The prompt for summarizing the transcript in segments is:

\begin{tcolorbox}[colback=gray!10, colframe=gray!50, sharp corners, boxrule=0.5mm]
\texttt{Given the following portion of a conversation, answer the question: \{query\}\\
The portion of the conversation is:\\
\{transcript\}}
\end{tcolorbox}

\noindent
and the segment summaries are then summarized into one final summary with:

\begin{tcolorbox}[colback=gray!10, colframe=gray!50, sharp corners, boxrule=0.5mm]
\texttt{The following is an ordered list of answers collected from portions of a conversation for the question: \{query\}\\
Generate a final answer for the question by aggregating the answers from the different conversation portions. Be succinct, and write it as a standalone answer without referring to the list of existing answers. The answers are:\\
Answer 1: \{answers[0]\}\\
Answer 2: \{answers[1]\}\\
...
}
\end{tcolorbox}

Notice that the prompt is phrased as if the query is a question and the summary is an answer. This layout is used to adhere to QMSum's format. The query for all the generic summaries is \textit{``Summarize the meeting''}. An example query for a query-focused summary is \textit{``What was the next step on features?''} or \textit{``Summarize what was said on intentionality''}.

The approximate cost for inferring on the QMSum test set with \texttt{gpt-4o-mini} was \$0.30.

\subsection{Question-answering with QAConv}
\label{sec_appendix_prompts_qaconv}

For question-answering, the following prompt is input to the LLM:

\begin{tcolorbox}[colback=gray!10, colframe=gray!50, sharp corners, boxrule=0.5mm]
\texttt{You will be given a conversation and some questions, and you need to answer the questions based on the conversation.\\
Each answer should be a very short span copied from the conversation, and written as a brief direct answer, and not as a sentence. Do not add any explanation or extra wording.\\
For example, for a question such as ``Where is John from?'', the answer could be ``New York'' but not ``John is from New York''.\\
If a question cannot be answered according to the conversation, answer with ``unanswerable'' only, without any explanation or extra wording.\\
Answer the questions line by line in the same order as the questions, without repeating the questions.\\\\
The conversation is:\\\{transcript\}\\\\
The questions are:\\
\{questions[0]\}\\
\{questions[1]\}\\
...}
\end{tcolorbox}

The transcript is chunked to fit within the context window of the LLM. Therefore, after answering the questions a chunk at a time, the final answer for a question is the shortest answer that is not ``unanswerable''. The default answer is ``unanswerable'' if no other answer is available.

The approximate cost for inferring on the QAConv test set with \texttt{gpt-4o-mini} was \$0.25.

\subsection{Dialog-act Classification with MRDA}
\label{sec_appendix_prompts_mrda}

The prompt for classifying dialog acts is:

\begin{tcolorbox}[colback=gray!10, colframe=gray!50, sharp corners, boxrule=0.5mm]
\small{
\texttt{Given an utterance from a conversation, choose a label that best describes the utterance.\\
The possible labels with their definitions are:\\
Floor Holder - the utterance occurs mid-speech and used by a speaker as a means to pause and continue holding the floor\\
Floor Grabber - an utterance in which a speaker has not been speaking and wants to gain the floor so that he may commence speaking\\
Hold Before Answer - an utterance that is used when a speaker who is given the floor and is expected to speak holds off prior to making an utterance\\
Agreement - an utterance used to exhibit agreement to or acceptance of a previous speaker's question, proposal, or statement\\
Yes-No-question - the utterance is in the form of a yes/no questions\\
Wh-Question - the utterance is a question that require a specific answer\\
Or-Clause - the utterance is an ``or'' clause, likely following a yes/no question\\
Or Question - the utterance offers the listener at least two answers or options from which to choose\\
Open-ended Question - the utterance is an open-ended question that places few syntactic or semantic constraints on the form of the answer it elicits\\
Rhetorical Question - the utterance states a question to which no answer is expected\\
Abandoned/Interrupted - an incomplete utterance in which a speaker stops talking intentionally or on account of being interrupted by another speaker\\
Uninterpretable - the utterance is not clear or has indecipherable speech\\
Continuer - the utterance is made in the background and simply indicate that a listener is following along or at least is yielding the illusion that he is paying attention\\
Statement - the utterance is none of the above types\\\\
The utterance is:\\
\{transcript\_utterance\}\\\\
The output should be in the format:\\
label: <the label>}}
\end{tcolorbox}

The approximate cost for inferring on the MRDA test set with \texttt{gpt-4o-mini} was \$0.10.

\section{Compute and Hardware}
\label{sec_appendix_compute}
For the components of the pipeline that require it, we use a single Nvidia A100 GPU with 40GB memory. This is needed for running Tortoise TTS, Whisper STT and for running open-source task models (Mistral and Llama models). The rest of the components run on an Apple M1 Macbook.

\section{More Results from Experiments}
\label{sec_appendix_results}
The graphs presented in Section \ref{sec_results} only show results when using pairwise comparison for summarization evaluation, fuzzy match for QA evaluation, and macro-$F_1$ for dialog act classification evaluation. Here we present the results also for the rest of the evaluation metrics for the three tasks.

\paragraph{Comparing task models.}
\autoref{fig_noclean_graphs_all} presents the graphs based on the rest of the task evaluation metrics, including those already presented in \autoref{fig_noclean_graphs}. \autoref{tab_scores_noclean_all} places the AUC and noise-toleration point scores for each curve in a single table for readability. The highest AUC score in each row is in bold, to show the task model achieving the best result. Significance can be inferred with the margin-of-errors.

For summarization, we find that the AUC scores do not differ significantly across models in all metrics, with the exception of Llama-3.1 having the highest AUC when using ROUGE-2 as the metric. The ROUGE metrics in general produce large margin-of-errors, causing comparison between models to be more vague. We do however notice that Llama-3.1 yields a larger difference in results as WER increases, when using all summarization metrics. We also see a consistent trend in Llama-3 where there seems to be a sudden drop and gain around a WER of 0.2.

The three metrics used for QAConv are quite consistent. For MRDA, we find that the accuracy metric raises the ranking of the Mistral model, since Mistral was always more likely to output a label of a prevalent class, stressing the advantage of the macro-$F_1$ metric that balances the importance of the classes.

For MRDA, we compare in Section \ref{sec_results} results to those in \citet{miah-etal-2023-hierarchical}.
They use an even more fine-grained label set (53 vs. 12 labels), yet still produce better scores overall. This further strengthens our presupposition that the models in our experiments are not as effective on the task, resulting in high noise-toleration-points on the curves.

\paragraph{Comparing noise types.}
Figures \ref{fig_cleaning_graphs_all_qmsum}, \ref{fig_cleaning_graphs_all_qaconv} and \ref{fig_cleaning_graphs_all_mrda} show the graphs based on the rest of the task models, evaluation metrics, and cleaning techniques, including those in \autoref{fig_cleaning_graphs}. \autoref{tab_scores_cleaning_all} shows all the cleaning effectiveness scores accordingly, similar to \autoref{tab_scores_cleaning}, but with the cleaning techniques kept in consistent order.

When looking at the cleaning-effectiveness scores, we find that ROUGE-2 ranks the cleaning techniques closest to the way that pairwise ranking does. In QAConv, the $F_1$ metric ranks the cleaning techniques very similarly for all four models. With the exception of Llama-3, the rankings are similar also with the exact and fuzzy matching metrics as well. For MRDA, although the score values are quite different, the rankings are close when using the two evaluation metrics.

As discussed in Section \ref{sec_results}, generally the named-entities and nouns are most helpful for the summarization and question-answering tasks. For GPT, this is also the case for dialog-act classification, however with the open-source models, the non-content words are most helpful for the task.

\input{tables/scores_noclean_all}

\input{tables/scores_cleaning_all}

\input{figures/noclean/noclean_main_all}

\input{figures/cleaning/cleaning_main_all_qmsum}

\input{figures/cleaning/cleaning_main_all_qaconv}

\input{figures/cleaning/cleaning_main_all_mrda}

\section{Task Datasets and Example Instances}
\label{sec_appendix_datasets}

\paragraph{QMSum.}
An example of an instance for summarization from the QMSum dataset is in \autoref{fig_qmsum_example}.

\paragraph{QAConv.}
An example of an instance for question-answering from the QAConv dataset is in \autoref{fig_qaconv_example}.

\paragraph{MRDA.}
Examples of instances for dialog act classification from the \textbf{MRDA} dataset are in \autoref{fig_mrda_example}.

The label-sets in dialog-act classification vary from one dataset to another, and have different levels of granularity (from 5 to 50+ labels). The utterances in MRDA are labeled with a dialog act on three granularity levels (with tag-sets of 5, 12 or 53 labels). For our experiments, we used the middle granularity level \citep[tagset with 12 dialog acts;][]{dhillon2004mrdaLabeling}, and the first and last 50 utterances from each transcript (100 of $\sim$1392), for efficiency purposes.

\input{figures/datasets/qmsum_example}

\input{figures/datasets/qaconv_example}

\input{figures/datasets/mrda_example}

\section{Licenses}

The following are the licenses of the used resources:
\begin{itemize}[noitemsep, topsep=0pt]
    \item Corpora:
    \begin{itemize}[noitemsep, topsep=0pt]
        \item QMSum: MIT
        \item QAConv: BSD-3-Clause (Salesforce)
        \item MRDA: GPL-3.0
    \end{itemize}
    \item Tools:
    \begin{itemize}[noitemsep, topsep=0pt]
        \item tortoise-tts: Apache-2.0
        \item rir-generator: MIT
        \item jiwer: Apache-2.0
        \item spacy: MIT
    \end{itemize}
    \item LLMs: Mistral and Llama models are gated models pulled from Huggingface.
\end{itemize}

\noindent
We use the above resources for exemplifying our framework, and solely for research purposes. Generally, our framework is intended for assessment of SLU solutions.

The code for our framework and analyses will be released under the Apache-2.0 license.

\section{Use of AI for the Paper}

ChatGPT was used for some minor rephrasing of sentences within the paper, and for assisting in preparing code to programmatically fill tables in LaTeX (placing results in tables for the paper).

%% file: figures/graph_illustration.tex
\begin{figure}[ht]
    \centering
    \includegraphics[width=\linewidth, trim=590 225 118 101, clip]{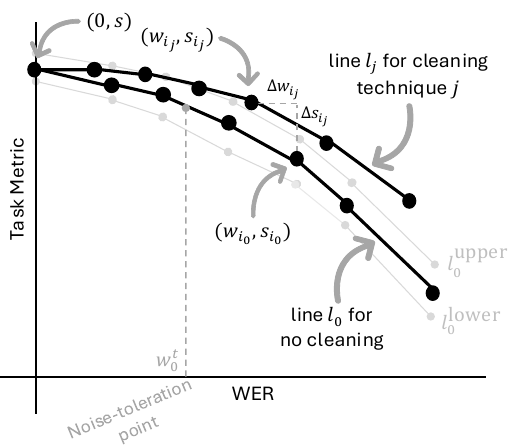}
    \caption{An illustration of the graph generated with the framework, for visual reference.}
    \label{fig_framework_graph}
\end{figure}

%% file: tables/scores_noclean_all.tex
\begin{table*}[ht]
    \centering
    \resizebox{\textwidth}{!}{%
    \begin{tabular}{c|l|cc|cc|cc|cc}
    \toprule
 & & \multicolumn{2}{c|}{Mistral-7B} & \multicolumn{2}{c|}{Llama3-8B} & \multicolumn{2}{c|}{Llama3.1-8B} & \multicolumn{2}{c}{GPT-4o-mini} \\
    & Metric & AUC & NTP & AUC & NTP & AUC & NTP & AUC & NTP \\
    \midrule
    \small{\multirow{4}{*}{\rotatebox[origin=c]{90}{QMSum}}} &
    PW-Rank &     \textbf{5.817} \small{± 0.321} & 0.195 &     5.811 \small{± 0.331} & 0.070 &     5.746 \small{± 0.312} & 0.249 &     5.627 \small{± 0.280} & 0.297 \\
    &
    R-1 &     0.227 \small{± 0.008} & - &     \textbf{0.228} \small{± 0.008} & 0.487 &     0.210 \small{± 0.010} & 0.526 &     0.225 \small{± 0.008} & - \\
    &
    R-2 &     0.051 \small{± 0.004} & 0.631 &     0.048 \small{± 0.004} & 0.460 &     \textbf{0.062} \small{± 0.006} & 0.494 &     0.051 \small{± 0.004} & 0.598 \\
    &
    R-L &     0.143 \small{± 0.005} & - &     \textbf{0.147} \small{± 0.005} & 0.586 &     0.140 \small{± 0.007} & 0.517 &     0.138 \small{± 0.005} & - \\
    \midrule
    \small{\multirow{3}{*}{\rotatebox[origin=c]{90}{QAConv}}} &
    Fuzzy &     39.080 \small{± 1.241} & 0.341 &     37.331 \small{± 1.217} & 0.302 &     49.466 \small{± 1.288} & 0.051 &     \textbf{49.992} \small{± 1.286} & 0.214 \\
    &
    $F_1$ &     0.289 \small{± 0.015} & 0.324 &     0.255 \small{± 0.014} & 0.269 &     0.384 \small{± 0.016} & 0.037 &     \textbf{0.392} \small{± 0.016} & 0.050 \\
    &
    Exact &     0.179 \small{± 0.015} & 0.293 &     0.164 \small{± 0.014} & 0.273 &     0.277 \small{± 0.017} & 0.031 &     \textbf{0.291} \small{± 0.017} & 0.042 \\
    \midrule
    \small{\multirow{2}{*}{\rotatebox[origin=c]{90}{MRDA}}} &
    Mac-$F_1$ &     0.110 \small{± 0.019} & 0.560 &     0.149 \small{± 0.028} & 0.615 &     0.140 \small{± 0.028} & 0.709 &     \textbf{0.193} \small{± 0.032} & 0.439 \\
    &
    Acc &     0.331 \small{± 0.023} & - &     0.263 \small{± 0.023} & 0.841 &     0.253 \small{± 0.022} & 0.790 &     \textbf{0.371} \small{± 0.024} & 0.553 \\
    \bottomrule
    \end{tabular}%
    }
    \caption{The area-under-the-curve (AUC; with margin of error at 95\% confidence level) and the noise-toleration point (NTP) of the experimented task models on the three tasks (summarization, question-answering and dialog-act classification). The highest AUC in each row is in bold. The scores correspond to the graphs in \autoref{fig_noclean_graphs_all}, where a more in-depth examination can be conducted.}
    \label{tab_scores_noclean_all}
\end{table*}

%% file: tables/scores_cleaning_all.tex
\begin{table*}[ht]
    \centering
    \resizebox{0.87\columnwidth}{!}{%
    \begin{tabular}{c|lc|c|c|c}
    \toprule
 & & Mistral & Llama3 & Llama3.1 & GPT4oMini \\
    \midrule
    \multirow{7}{*}{\rotatebox[origin=c]{90}{QMSum | PW-Rank}}
    &
    Adjs. &     \cellcolor[gray]{0.67} \textcolor{black}{0.327} &     \cellcolor[gray]{0.83} \textcolor{black}{1.216} &     \cellcolor[gray]{0.67} \textcolor{black}{0.230} &     \cellcolor[gray]{0.33} \textcolor{white}{0.135} \\
    &
    Advs. &     \cellcolor[gray]{0.00} \textcolor{white}{0.223} &     \cellcolor[gray]{0.67} \textcolor{black}{0.956} &     \cellcolor[gray]{0.00} \textcolor{white}{0.017} &     \cellcolor[gray]{0.00} \textcolor{white}{-0.023} \\
    &
    Content &     \cellcolor[gray]{0.50} \textcolor{white}{0.322} &     \cellcolor[gray]{0.17} \textcolor{white}{0.581} &     \cellcolor[gray]{0.83} \textcolor{black}{0.357} &     \cellcolor[gray]{0.83} \textcolor{black}{0.479} \\
    &
    N-ents. &     \cellcolor[gray]{1.00} \textcolor{black}{0.537} &     \cellcolor[gray]{1.00} \textcolor{black}{1.346} &     \cellcolor[gray]{0.33} \textcolor{white}{0.193} &     \cellcolor[gray]{1.00} \textcolor{black}{0.499} \\
    &
    Non-cont, &     \cellcolor[gray]{0.33} \textcolor{white}{0.262} &     \cellcolor[gray]{0.00} \textcolor{white}{0.480} &     \cellcolor[gray]{0.50} \textcolor{white}{0.209} &     \cellcolor[gray]{0.50} \textcolor{white}{0.181} \\
    &
    Nouns &     \cellcolor[gray]{0.83} \textcolor{black}{0.459} &     \cellcolor[gray]{0.50} \textcolor{white}{0.804} &     \cellcolor[gray]{1.00} \textcolor{black}{0.384} &     \cellcolor[gray]{0.67} \textcolor{black}{0.305} \\
    &
    Verbs &     \cellcolor[gray]{0.17} \textcolor{white}{0.229} &     \cellcolor[gray]{0.33} \textcolor{white}{0.702} &     \cellcolor[gray]{0.17} \textcolor{white}{0.145} &     \cellcolor[gray]{0.17} \textcolor{white}{0.073} \\
    \cmidrule(lr){2-6}
    \multirow{7}{*}{\rotatebox[origin=c]{90}{QMSum | R-1}}
    &
    Adjs. &     \cellcolor[gray]{0.50} \textcolor{white}{-0.068} &     \cellcolor[gray]{1.00} \textcolor{black}{0.237} &     \cellcolor[gray]{0.17} \textcolor{white}{-0.038} &     \cellcolor[gray]{1.00} \textcolor{black}{0.071} \\
    &
    Advs. &     \cellcolor[gray]{0.33} \textcolor{white}{-0.071} &     \cellcolor[gray]{0.83} \textcolor{black}{0.203} &     \cellcolor[gray]{0.00} \textcolor{white}{-0.058} &     \cellcolor[gray]{0.67} \textcolor{black}{0.059} \\
    &
    Content &     \cellcolor[gray]{1.00} \textcolor{black}{-0.006} &     \cellcolor[gray]{0.33} \textcolor{white}{0.119} &     \cellcolor[gray]{1.00} \textcolor{black}{0.078} &     \cellcolor[gray]{0.17} \textcolor{white}{-0.003} \\
    &
    N-ents. &     \cellcolor[gray]{0.00} \textcolor{white}{-0.085} &     \cellcolor[gray]{0.67} \textcolor{black}{0.198} &     \cellcolor[gray]{0.33} \textcolor{white}{-0.005} &     \cellcolor[gray]{0.00} \textcolor{white}{-0.115} \\
    &
    Non-cont, &     \cellcolor[gray]{0.83} \textcolor{black}{-0.028} &     \cellcolor[gray]{0.00} \textcolor{white}{0.090} &     \cellcolor[gray]{0.67} \textcolor{black}{0.043} &     \cellcolor[gray]{0.33} \textcolor{white}{0.033} \\
    &
    Nouns &     \cellcolor[gray]{0.67} \textcolor{black}{-0.038} &     \cellcolor[gray]{0.50} \textcolor{white}{0.145} &     \cellcolor[gray]{0.83} \textcolor{black}{0.058} &     \cellcolor[gray]{0.50} \textcolor{white}{0.057} \\
    &
    Verbs &     \cellcolor[gray]{0.17} \textcolor{white}{-0.074} &     \cellcolor[gray]{0.17} \textcolor{white}{0.103} &     \cellcolor[gray]{0.50} \textcolor{white}{-0.003} &     \cellcolor[gray]{0.83} \textcolor{black}{0.062} \\
    \cmidrule(lr){2-6}
    \multirow{7}{*}{\rotatebox[origin=c]{90}{QMSum | R-2}}
    &
    Adjs. &     \cellcolor[gray]{0.50} \textcolor{white}{0.000} &     \cellcolor[gray]{0.67} \textcolor{black}{0.403} &     \cellcolor[gray]{0.17} \textcolor{white}{0.054} &     \cellcolor[gray]{0.67} \textcolor{black}{0.198} \\
    &
    Advs. &     \cellcolor[gray]{0.33} \textcolor{white}{-0.084} &     \cellcolor[gray]{0.83} \textcolor{black}{0.422} &     \cellcolor[gray]{0.00} \textcolor{white}{0.011} &     \cellcolor[gray]{0.17} \textcolor{white}{0.134} \\
    &
    Content &     \cellcolor[gray]{1.00} \textcolor{black}{0.053} &     \cellcolor[gray]{0.33} \textcolor{white}{0.234} &     \cellcolor[gray]{0.83} \textcolor{black}{0.212} &     \cellcolor[gray]{0.83} \textcolor{black}{0.249} \\
    &
    N-ents. &     \cellcolor[gray]{0.17} \textcolor{white}{-0.093} &     \cellcolor[gray]{1.00} \textcolor{black}{0.492} &     \cellcolor[gray]{0.67} \textcolor{black}{0.140} &     \cellcolor[gray]{1.00} \textcolor{black}{0.265} \\
    &
    Non-cont, &     \cellcolor[gray]{0.67} \textcolor{black}{0.003} &     \cellcolor[gray]{0.00} \textcolor{white}{0.188} &     \cellcolor[gray]{0.50} \textcolor{white}{0.104} &     \cellcolor[gray]{0.00} \textcolor{white}{0.101} \\
    &
    Nouns &     \cellcolor[gray]{0.83} \textcolor{black}{0.028} &     \cellcolor[gray]{0.50} \textcolor{white}{0.327} &     \cellcolor[gray]{1.00} \textcolor{black}{0.252} &     \cellcolor[gray]{0.50} \textcolor{white}{0.189} \\
    &
    Verbs &     \cellcolor[gray]{0.00} \textcolor{white}{-0.131} &     \cellcolor[gray]{0.17} \textcolor{white}{0.208} &     \cellcolor[gray]{0.33} \textcolor{white}{0.066} &     \cellcolor[gray]{0.33} \textcolor{white}{0.156} \\
    \cmidrule(lr){2-6}
    \multirow{7}{*}{\rotatebox[origin=c]{90}{QMSum | R-L}}
    &
    Adjs. &     \cellcolor[gray]{0.33} \textcolor{white}{-0.092} &     \cellcolor[gray]{1.00} \textcolor{black}{0.189} &     \cellcolor[gray]{0.00} \textcolor{white}{-0.015} &     \cellcolor[gray]{0.67} \textcolor{black}{0.050} \\
    &
    Advs. &     \cellcolor[gray]{0.17} \textcolor{white}{-0.100} &     \cellcolor[gray]{0.67} \textcolor{black}{0.136} &     \cellcolor[gray]{0.17} \textcolor{white}{-0.014} &     \cellcolor[gray]{0.83} \textcolor{black}{0.054} \\
    &
    Content &     \cellcolor[gray]{1.00} \textcolor{black}{-0.022} &     \cellcolor[gray]{0.33} \textcolor{white}{0.083} &     \cellcolor[gray]{0.83} \textcolor{black}{0.081} &     \cellcolor[gray]{0.17} \textcolor{white}{0.008} \\
    &
    N-ents. &     \cellcolor[gray]{0.00} \textcolor{white}{-0.120} &     \cellcolor[gray]{0.83} \textcolor{black}{0.144} &     \cellcolor[gray]{0.33} \textcolor{white}{0.014} &     \cellcolor[gray]{0.00} \textcolor{white}{-0.052} \\
    &
    Non-cont, &     \cellcolor[gray]{0.83} \textcolor{black}{-0.038} &     \cellcolor[gray]{0.00} \textcolor{white}{0.061} &     \cellcolor[gray]{0.67} \textcolor{black}{0.049} &     \cellcolor[gray]{0.33} \textcolor{white}{0.018} \\
    &
    Nouns &     \cellcolor[gray]{0.67} \textcolor{black}{-0.040} &     \cellcolor[gray]{0.50} \textcolor{white}{0.120} &     \cellcolor[gray]{1.00} \textcolor{black}{0.088} &     \cellcolor[gray]{0.50} \textcolor{white}{0.041} \\
    &
    Verbs &     \cellcolor[gray]{0.50} \textcolor{white}{-0.081} &     \cellcolor[gray]{0.17} \textcolor{white}{0.065} &     \cellcolor[gray]{0.50} \textcolor{white}{0.031} &     \cellcolor[gray]{1.00} \textcolor{black}{0.057} \\
    \midrule
    \multirow{7}{*}{\rotatebox[origin=c]{90}{QAConv | Fuzzy}}
    &
    Adjs. &     \cellcolor[gray]{0.00} \textcolor{white}{-0.037} &     \cellcolor[gray]{0.83} \textcolor{black}{0.398} &     \cellcolor[gray]{0.50} \textcolor{white}{0.109} &     \cellcolor[gray]{0.33} \textcolor{white}{0.120} \\
    &
    Advs. &     \cellcolor[gray]{0.33} \textcolor{white}{0.004} &     \cellcolor[gray]{0.67} \textcolor{black}{0.335} &     \cellcolor[gray]{0.00} \textcolor{white}{0.033} &     \cellcolor[gray]{0.00} \textcolor{white}{0.071} \\
    &
    Content &     \cellcolor[gray]{0.67} \textcolor{black}{0.108} &     \cellcolor[gray]{0.33} \textcolor{white}{0.224} &     \cellcolor[gray]{0.67} \textcolor{black}{0.186} &     \cellcolor[gray]{0.67} \textcolor{black}{0.202} \\
    &
    N-ents. &     \cellcolor[gray]{1.00} \textcolor{black}{0.221} &     \cellcolor[gray]{1.00} \textcolor{black}{0.469} &     \cellcolor[gray]{1.00} \textcolor{black}{0.294} &     \cellcolor[gray]{1.00} \textcolor{black}{0.311} \\
    &
    Non-cont, &     \cellcolor[gray]{0.50} \textcolor{white}{0.070} &     \cellcolor[gray]{0.00} \textcolor{white}{0.177} &     \cellcolor[gray]{0.33} \textcolor{white}{0.094} &     \cellcolor[gray]{0.50} \textcolor{white}{0.133} \\
    &
    Nouns &     \cellcolor[gray]{0.83} \textcolor{black}{0.164} &     \cellcolor[gray]{0.50} \textcolor{white}{0.280} &     \cellcolor[gray]{0.83} \textcolor{black}{0.210} &     \cellcolor[gray]{0.83} \textcolor{black}{0.211} \\
    &
    Verbs &     \cellcolor[gray]{0.17} \textcolor{white}{-0.012} &     \cellcolor[gray]{0.17} \textcolor{white}{0.188} &     \cellcolor[gray]{0.17} \textcolor{white}{0.059} &     \cellcolor[gray]{0.17} \textcolor{white}{0.090} \\
    \cmidrule(lr){2-6}
    \multirow{7}{*}{\rotatebox[origin=c]{90}{QAConv | $F_1$}}
    &
    Adjs. &     \cellcolor[gray]{0.00} \textcolor{white}{-0.376} &     \cellcolor[gray]{0.00} \textcolor{white}{-0.050} &     \cellcolor[gray]{0.33} \textcolor{white}{0.175} &     \cellcolor[gray]{0.33} \textcolor{white}{0.145} \\
    &
    Advs. &     \cellcolor[gray]{0.17} \textcolor{white}{-0.276} &     \cellcolor[gray]{0.17} \textcolor{white}{0.063} &     \cellcolor[gray]{0.00} \textcolor{white}{0.020} &     \cellcolor[gray]{0.00} \textcolor{white}{0.102} \\
    &
    Content &     \cellcolor[gray]{0.67} \textcolor{black}{0.152} &     \cellcolor[gray]{0.67} \textcolor{black}{0.304} &     \cellcolor[gray]{0.67} \textcolor{black}{0.318} &     \cellcolor[gray]{0.67} \textcolor{black}{0.325} \\
    &
    N-ents. &     \cellcolor[gray]{1.00} \textcolor{black}{0.271} &     \cellcolor[gray]{1.00} \textcolor{black}{0.584} &     \cellcolor[gray]{1.00} \textcolor{black}{0.510} &     \cellcolor[gray]{1.00} \textcolor{black}{0.493} \\
    &
    Non-cont, &     \cellcolor[gray]{0.50} \textcolor{white}{0.070} &     \cellcolor[gray]{0.50} \textcolor{white}{0.204} &     \cellcolor[gray]{0.50} \textcolor{white}{0.192} &     \cellcolor[gray]{0.50} \textcolor{white}{0.227} \\
    &
    Nouns &     \cellcolor[gray]{0.83} \textcolor{black}{0.203} &     \cellcolor[gray]{0.83} \textcolor{black}{0.372} &     \cellcolor[gray]{0.83} \textcolor{black}{0.378} &     \cellcolor[gray]{0.83} \textcolor{black}{0.347} \\
    &
    Verbs &     \cellcolor[gray]{0.33} \textcolor{white}{-0.095} &     \cellcolor[gray]{0.33} \textcolor{white}{0.085} &     \cellcolor[gray]{0.17} \textcolor{white}{0.110} &     \cellcolor[gray]{0.17} \textcolor{white}{0.134} \\
    \cmidrule(lr){2-6}
    \multirow{7}{*}{\rotatebox[origin=c]{90}{QAConv | Exact}}
    &
    Adjs. &     \cellcolor[gray]{0.17} \textcolor{white}{-0.003} &     \cellcolor[gray]{0.83} \textcolor{black}{0.948} &     \cellcolor[gray]{0.33} \textcolor{white}{0.221} &     \cellcolor[gray]{0.33} \textcolor{white}{0.225} \\
    &
    Advs. &     \cellcolor[gray]{0.00} \textcolor{white}{-0.049} &     \cellcolor[gray]{0.67} \textcolor{black}{0.852} &     \cellcolor[gray]{0.00} \textcolor{white}{0.014} &     \cellcolor[gray]{0.00} \textcolor{white}{0.141} \\
    &
    Content &     \cellcolor[gray]{0.67} \textcolor{black}{0.329} &     \cellcolor[gray]{0.33} \textcolor{white}{0.573} &     \cellcolor[gray]{0.67} \textcolor{black}{0.391} &     \cellcolor[gray]{0.67} \textcolor{black}{0.399} \\
    &
    N-ents. &     \cellcolor[gray]{1.00} \textcolor{black}{0.657} &     \cellcolor[gray]{1.00} \textcolor{black}{1.162} &     \cellcolor[gray]{1.00} \textcolor{black}{0.692} &     \cellcolor[gray]{1.00} \textcolor{black}{0.688} \\
    &
    Non-cont, &     \cellcolor[gray]{0.50} \textcolor{white}{0.215} &     \cellcolor[gray]{0.17} \textcolor{white}{0.462} &     \cellcolor[gray]{0.50} \textcolor{white}{0.221} &     \cellcolor[gray]{0.50} \textcolor{white}{0.284} \\
    &
    Nouns &     \cellcolor[gray]{0.83} \textcolor{black}{0.423} &     \cellcolor[gray]{0.50} \textcolor{white}{0.728} &     \cellcolor[gray]{0.83} \textcolor{black}{0.477} &     \cellcolor[gray]{0.83} \textcolor{black}{0.438} \\
    &
    Verbs &     \cellcolor[gray]{0.33} \textcolor{white}{0.001} &     \cellcolor[gray]{0.00} \textcolor{white}{0.441} &     \cellcolor[gray]{0.17} \textcolor{white}{0.131} &     \cellcolor[gray]{0.17} \textcolor{white}{0.170} \\
    \midrule
    \multirow{7}{*}{\rotatebox[origin=c]{90}{MRDA | Mac-$F_1$}}
    &
    Adjs. &     \cellcolor[gray]{1.00} \textcolor{black}{0.404} &     \cellcolor[gray]{0.50} \textcolor{white}{-0.001} &     \cellcolor[gray]{0.83} \textcolor{black}{0.035} &     \cellcolor[gray]{0.83} \textcolor{black}{0.290} \\
    &
    Advs. &     \cellcolor[gray]{0.00} \textcolor{white}{-0.043} &     \cellcolor[gray]{0.00} \textcolor{white}{-0.085} &     \cellcolor[gray]{0.00} \textcolor{white}{-0.342} &     \cellcolor[gray]{0.00} \textcolor{white}{0.107} \\
    &
    Content &     \cellcolor[gray]{0.67} \textcolor{black}{0.315} &     \cellcolor[gray]{0.67} \textcolor{black}{0.027} &     \cellcolor[gray]{0.50} \textcolor{white}{-0.053} &     \cellcolor[gray]{0.50} \textcolor{white}{0.212} \\
    &
    N-ents. &     \cellcolor[gray]{0.17} \textcolor{white}{-0.035} &     \cellcolor[gray]{0.17} \textcolor{white}{-0.049} &     \cellcolor[gray]{0.17} \textcolor{white}{-0.291} &     \cellcolor[gray]{1.00} \textcolor{black}{0.735} \\
    &
    Non-cont, &     \cellcolor[gray]{0.83} \textcolor{black}{0.392} &     \cellcolor[gray]{1.00} \textcolor{black}{0.122} &     \cellcolor[gray]{1.00} \textcolor{black}{0.122} &     \cellcolor[gray]{0.67} \textcolor{black}{0.285} \\
    &
    Nouns &     \cellcolor[gray]{0.33} \textcolor{white}{0.010} &     \cellcolor[gray]{0.83} \textcolor{black}{0.102} &     \cellcolor[gray]{0.67} \textcolor{black}{-0.042} &     \cellcolor[gray]{0.33} \textcolor{white}{0.203} \\
    &
    Verbs &     \cellcolor[gray]{0.50} \textcolor{white}{0.044} &     \cellcolor[gray]{0.33} \textcolor{white}{-0.024} &     \cellcolor[gray]{0.33} \textcolor{white}{-0.232} &     \cellcolor[gray]{0.17} \textcolor{white}{0.158} \\
    \cmidrule(lr){2-6}
    \multirow{7}{*}{\rotatebox[origin=c]{90}{MRDA | Acc}}
    &
    Adjs. &     \cellcolor[gray]{1.00} \textcolor{black}{-0.335} &     \cellcolor[gray]{0.33} \textcolor{white}{-0.619} &     \cellcolor[gray]{0.67} \textcolor{black}{-0.366} &     \cellcolor[gray]{0.83} \textcolor{black}{0.174} \\
    &
    Advs. &     \cellcolor[gray]{0.00} \textcolor{white}{-1.160} &     \cellcolor[gray]{0.00} \textcolor{white}{-1.035} &     \cellcolor[gray]{0.00} \textcolor{white}{-0.887} &     \cellcolor[gray]{0.00} \textcolor{white}{-0.207} \\
    &
    Content &     \cellcolor[gray]{0.83} \textcolor{black}{-0.373} &     \cellcolor[gray]{1.00} \textcolor{black}{-0.362} &     \cellcolor[gray]{0.83} \textcolor{black}{-0.350} &     \cellcolor[gray]{0.33} \textcolor{white}{-0.082} \\
    &
    N-ents. &     \cellcolor[gray]{0.17} \textcolor{white}{-1.087} &     \cellcolor[gray]{0.17} \textcolor{white}{-0.797} &     \cellcolor[gray]{0.17} \textcolor{white}{-0.618} &     \cellcolor[gray]{1.00} \textcolor{black}{0.245} \\
    &
    Non-cont, &     \cellcolor[gray]{0.33} \textcolor{white}{-0.685} &     \cellcolor[gray]{0.83} \textcolor{black}{-0.431} &     \cellcolor[gray]{0.50} \textcolor{white}{-0.386} &     \cellcolor[gray]{0.50} \textcolor{white}{-0.081} \\
    &
    Nouns &     \cellcolor[gray]{0.67} \textcolor{black}{-0.594} &     \cellcolor[gray]{0.67} \textcolor{black}{-0.437} &     \cellcolor[gray]{1.00} \textcolor{black}{-0.314} &     \cellcolor[gray]{0.67} \textcolor{black}{-0.047} \\
    &
    Verbs &     \cellcolor[gray]{0.50} \textcolor{white}{-0.654} &     \cellcolor[gray]{0.50} \textcolor{white}{-0.584} &     \cellcolor[gray]{0.33} \textcolor{white}{-0.436} &     \cellcolor[gray]{0.17} \textcolor{white}{-0.179} \\
    \bottomrule
    \end{tabular}%
    }
    \caption{The cleaning-effectiveness scores (CES) of the experimented cleaning techniques on the four task-models. In the experimented techniques, a certain group of words is repaired, with respect to the reference transcripts, in so demonstrating the effect of varying noise types. The corresponding graphs are shown in Figures \ref{fig_cleaning_graphs_all_qmsum}, \ref{fig_cleaning_graphs_all_qaconv} and \ref{fig_cleaning_graphs_all_mrda}.}
    \label{tab_scores_cleaning_all}
\end{table*}

%% file: figures/noclean/noclean_main_all.tex
\begin{figure*}[ht]
    \centering
    \subfloat[QMSum with pairwise ranking]{
        \includegraphics[width=0.32\linewidth, trim=8 8 7 21, clip]{figures/noclean/qmsum_noclean_pairwise_ranking.pdf}
    }
    \subfloat[QMSum with ROUGE-1]{
        \includegraphics[width=0.32\linewidth, trim=8 8 7 21, clip]{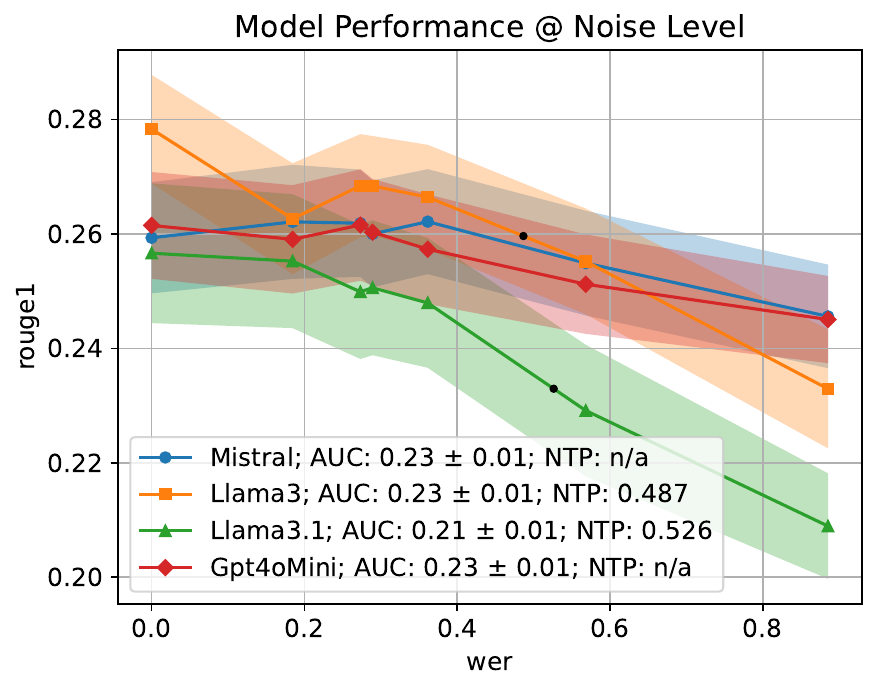}
    }
    \subfloat[QMSum with ROUGE-2]{
        \includegraphics[width=0.32\linewidth, trim=8 8 7 21, clip]{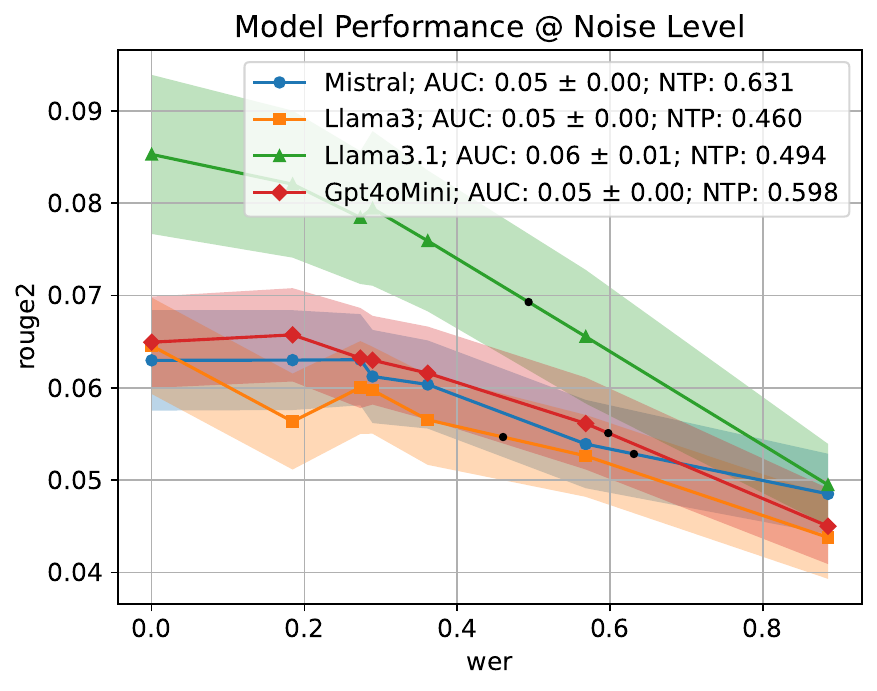}
    }
    \hspace{0.2cm}
    \subfloat[QMSum with ROUGE-L]{
        \includegraphics[width=0.32\linewidth, trim=8 8 7 21, clip]{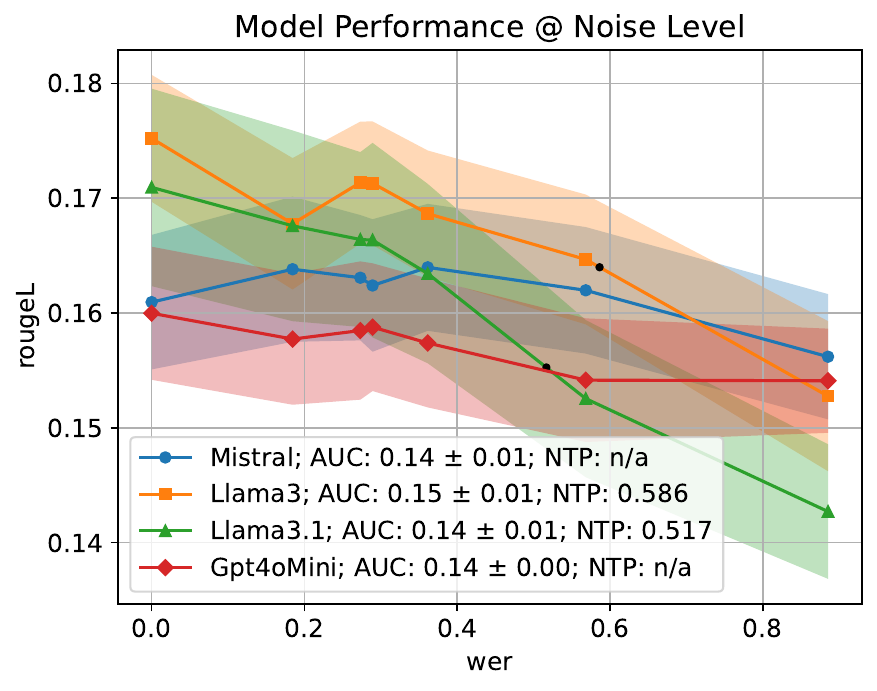}
    }
    \subfloat[MRDA with macro-$F_1$]{
        \includegraphics[width=0.32\linewidth, trim=8 8 7 21, clip]{figures/noclean/mrda_noclean_f1_macro.pdf}
    }
    \subfloat[MRDA with accuracy]{
        \includegraphics[width=0.32\linewidth, trim=8 8 7 21, clip]{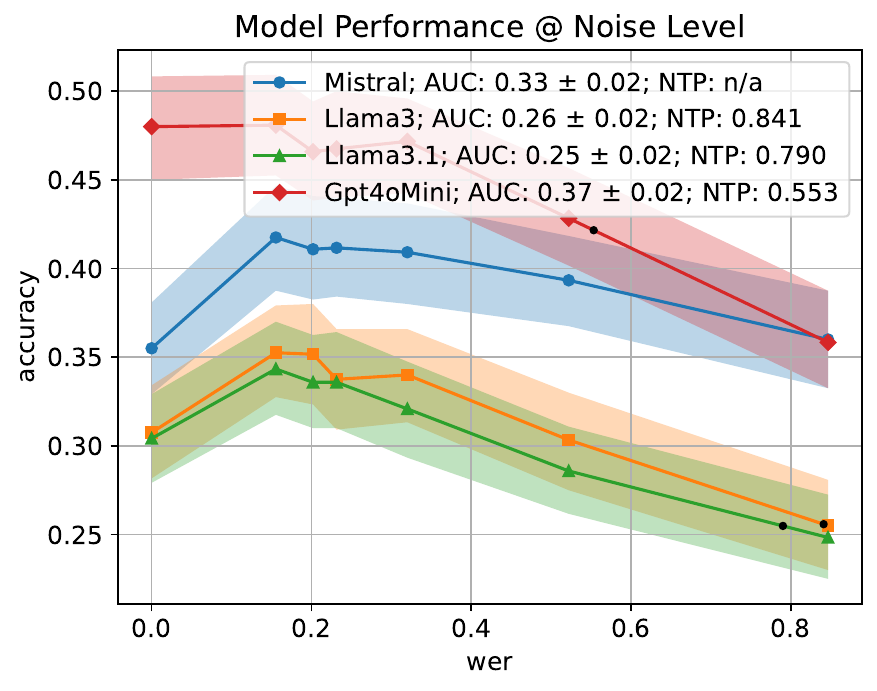}
    }
    \hspace{0.2cm}
    \subfloat[QAConv with fuzzy match]{
        \includegraphics[width=0.32\linewidth, trim=8 8 7 21, clip]{figures/noclean/qaconv_noclean_fzr.pdf}
    }
    \subfloat[QAConv with token $F_1$]{
        \includegraphics[width=0.32\linewidth, trim=8 8 7 21, clip]{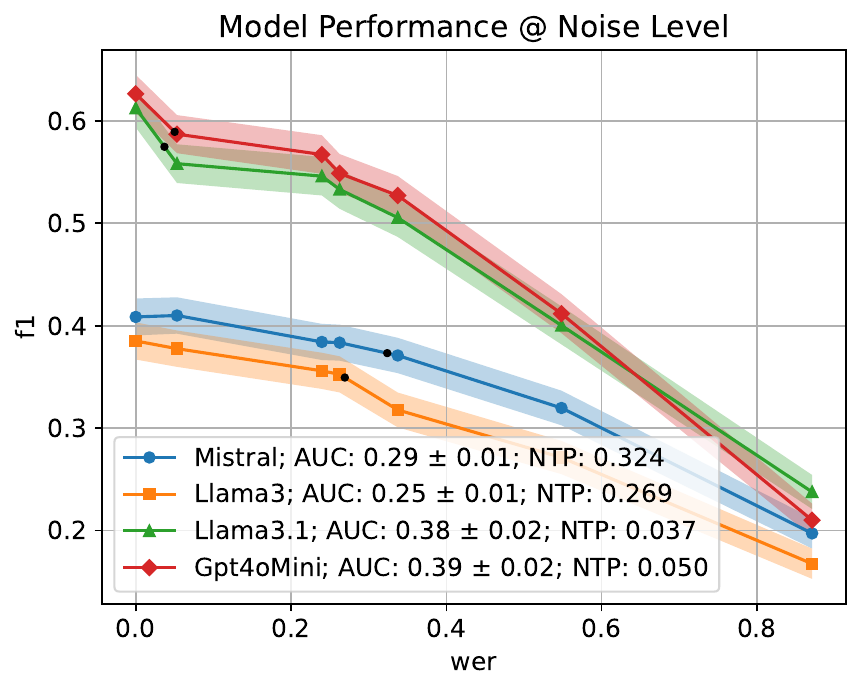}
    }
    \subfloat[QAConv with exact match]{
        \includegraphics[width=0.32\linewidth, trim=8 8 7 21, clip]{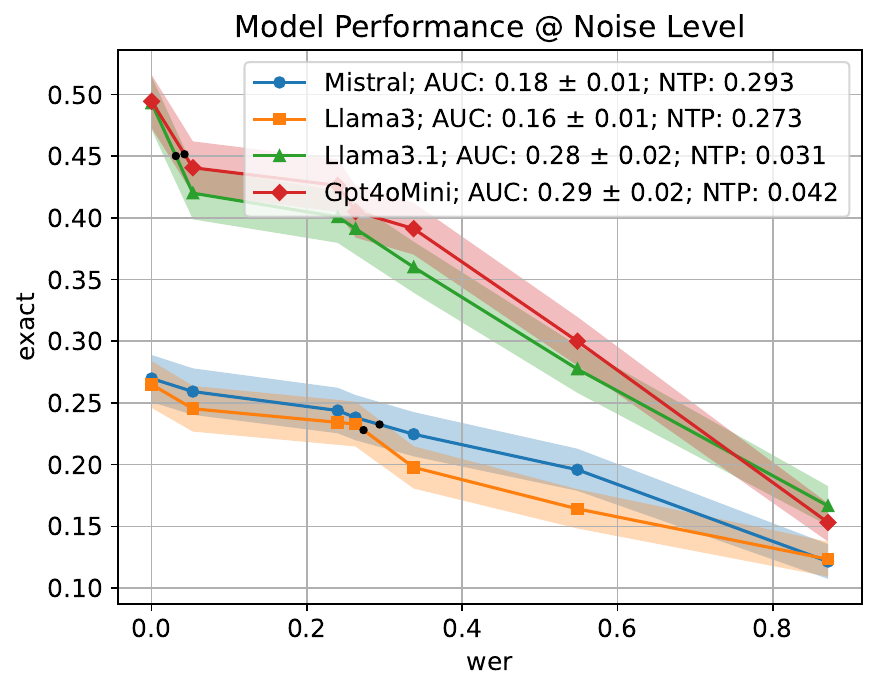}
    }
    \caption{The performance of the models on the tasks in our experiments. The curves provide insights into the performance of the models at various noise levels. The curves in each plot can be compared with their area-under-the-curve. Also, each curve is marked with its noise-toleration point, which provides the WER value where the task-score first decreases significantly, with respect to the score at $\text{WER}=0$. The shaded area around a line represents the corresponding confidence interval. The scores can be found in \autoref{tab_scores_noclean_all}.}
    \label{fig_noclean_graphs_all}
\end{figure*}

%% file: figures/cleaning/cleaning_main_all_qmsum.tex
\begin{figure*}[ht]
    \centering
    \subfloat[Pairwise ranking, Mistral]{
        \includegraphics[width=0.32\linewidth, trim=8 8 7 21, clip]{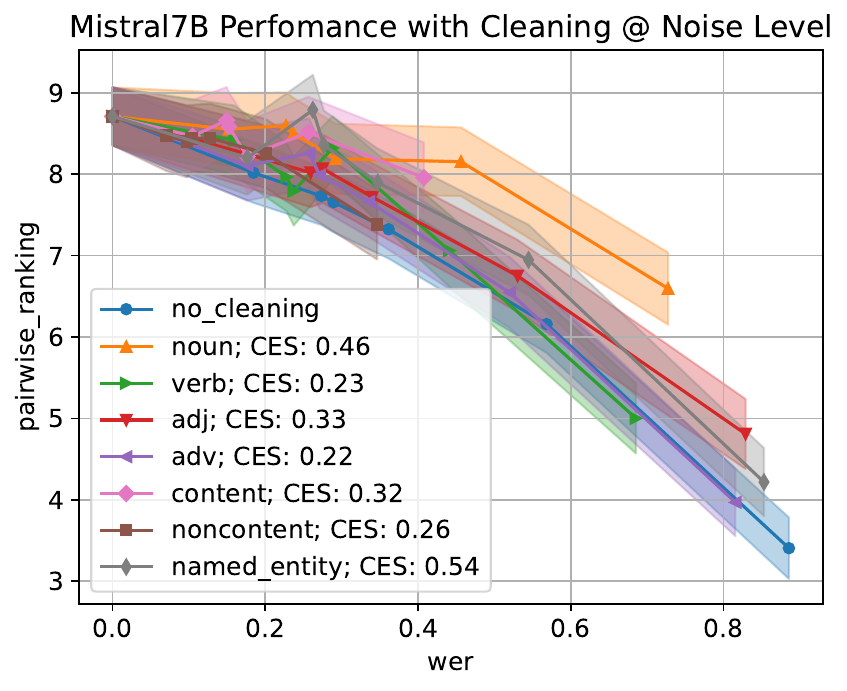}
    }
    \subfloat[ROUGE-1, Mistral]{
        \includegraphics[width=0.32\linewidth, trim=8 8 7 21, clip]{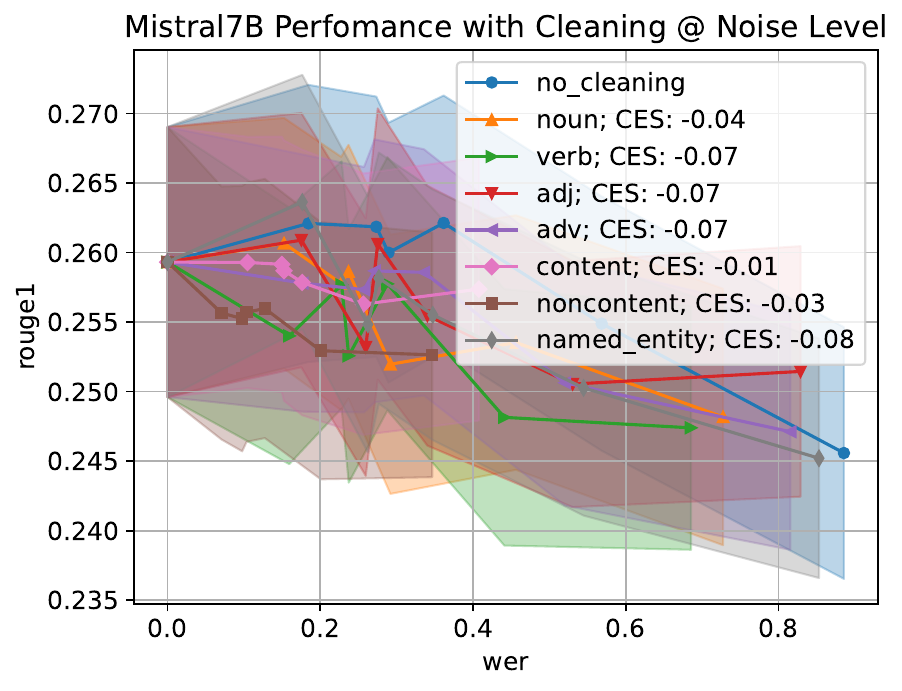}
    }
    \subfloat[ROUGE-2, Mistral]{
        \includegraphics[width=0.32\linewidth, trim=8 8 7 21, clip]{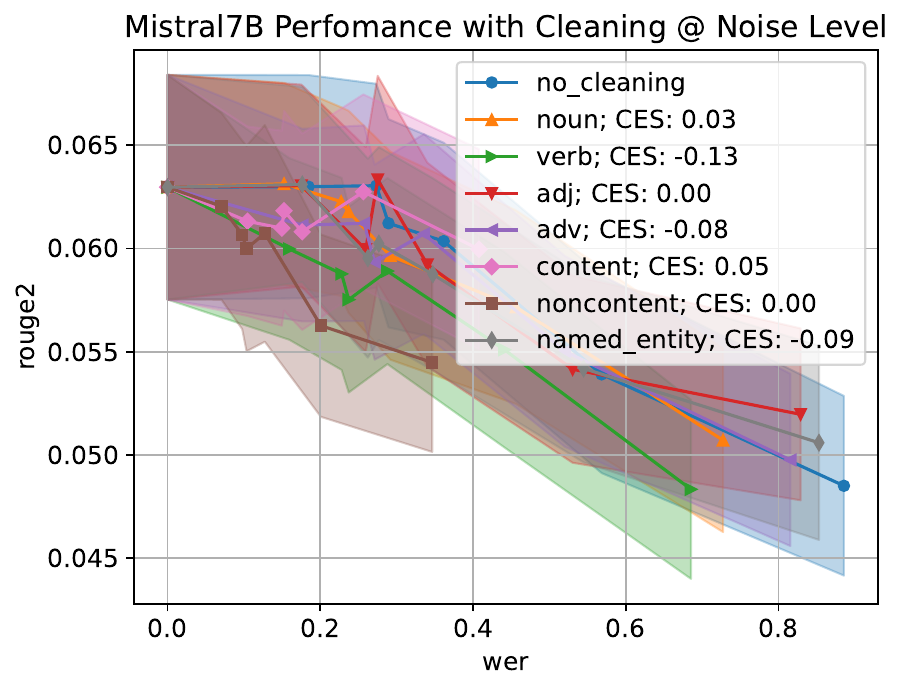}
    }
    \hspace{0.2cm}
    \subfloat[Pairwise ranking, Llama3]{
        \includegraphics[width=0.32\linewidth, trim=8 8 7 21, clip]{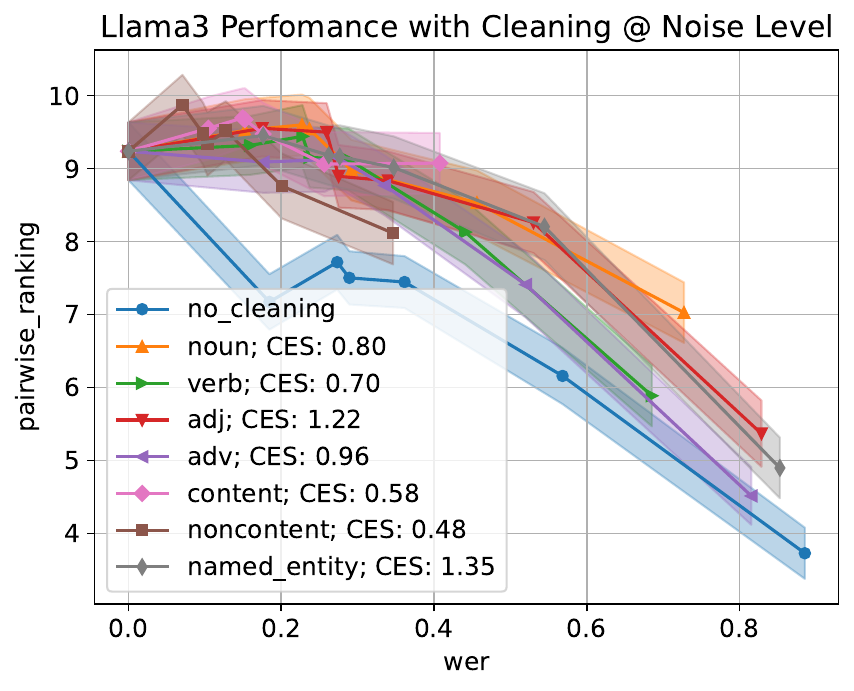}
    }
    \subfloat[ROUGE-1, Llama3]{
        \includegraphics[width=0.32\linewidth, trim=8 8 7 21, clip]{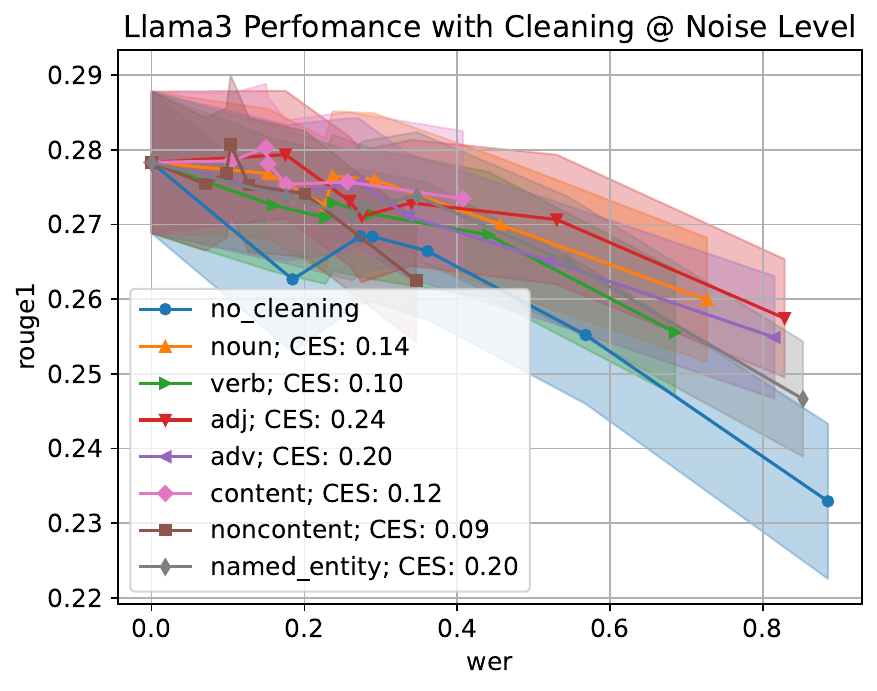}
    }
    \subfloat[ROUGE-2, Llama3]{
        \includegraphics[width=0.32\linewidth, trim=8 8 7 21, clip]{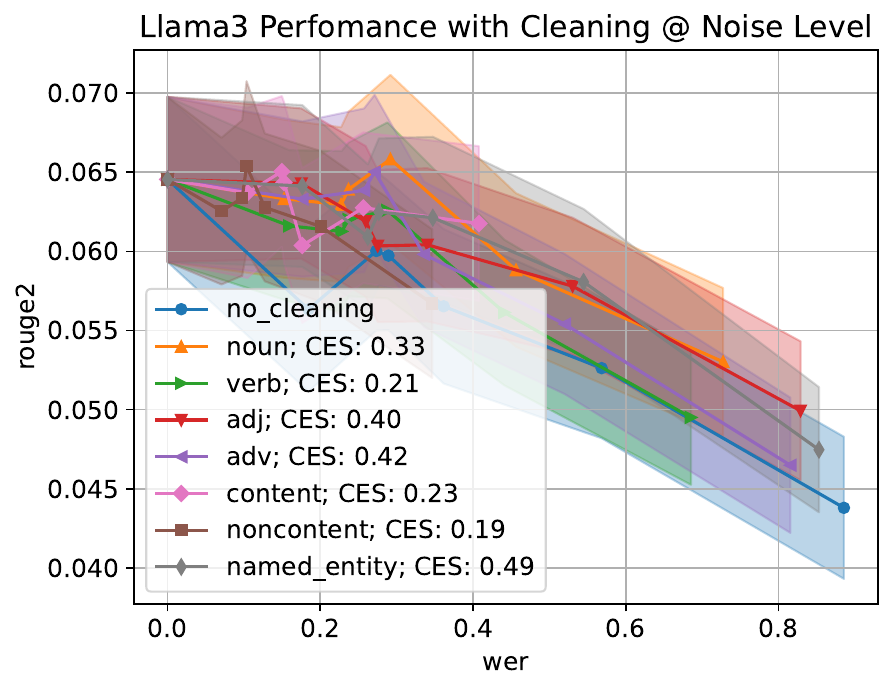}
    }
    \hspace{0.2cm}
    \subfloat[Pairwise ranking, Llama3.1]{
        \includegraphics[width=0.32\linewidth, trim=8 8 7 21, clip]{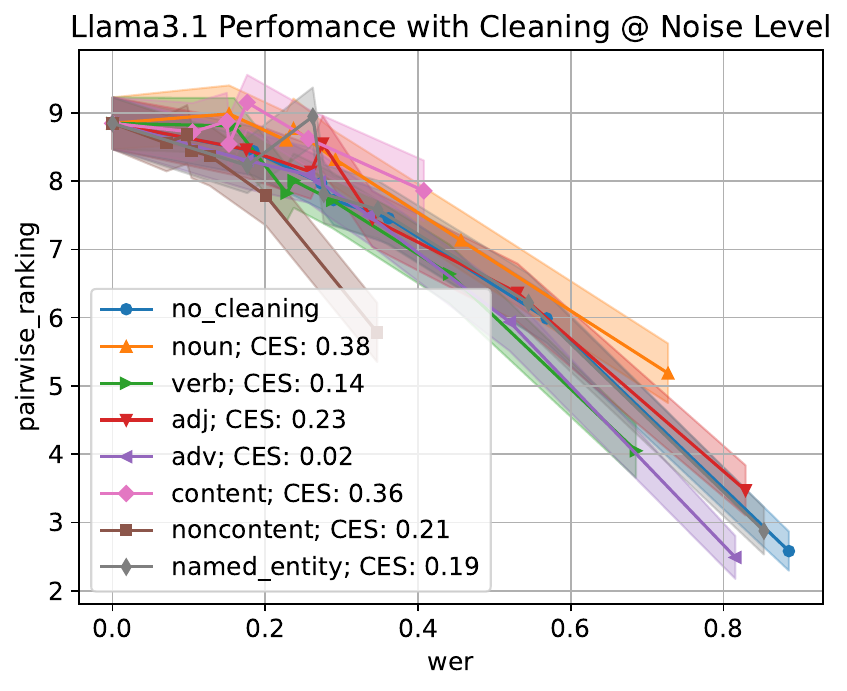}
    }
    \subfloat[ROUGE-1, Llama3.1]{
        \includegraphics[width=0.32\linewidth, trim=8 8 7 21, clip]{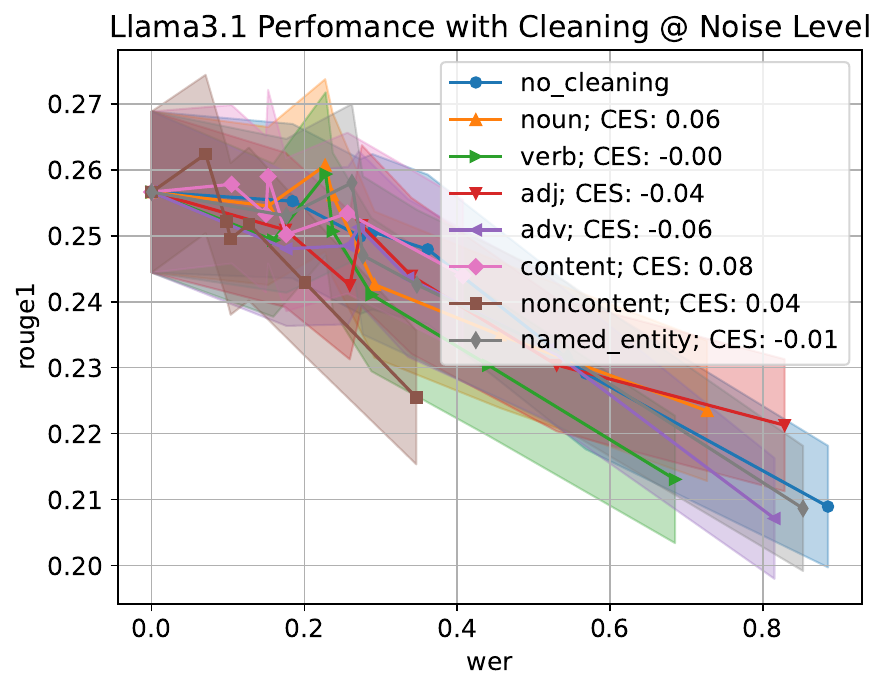}
    }
    \subfloat[ROUGE-2, Llama3.1]{
        \includegraphics[width=0.32\linewidth, trim=8 8 7 21, clip]{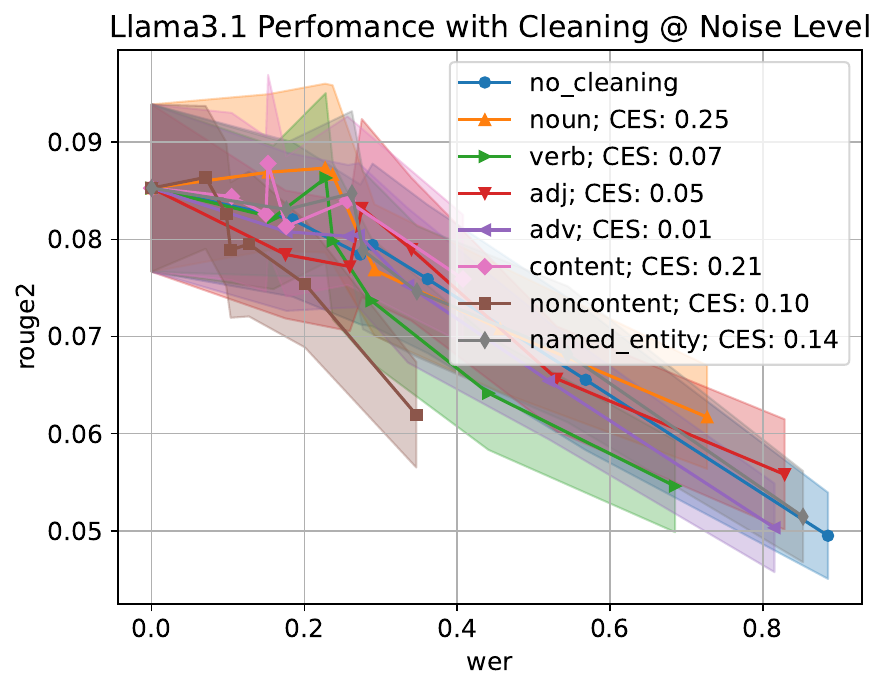}
    }
    \hspace{0.2cm}
    \subfloat[Pairwise ranking, GPT]{
        \includegraphics[width=0.32\linewidth, trim=8 8 7 21, clip]{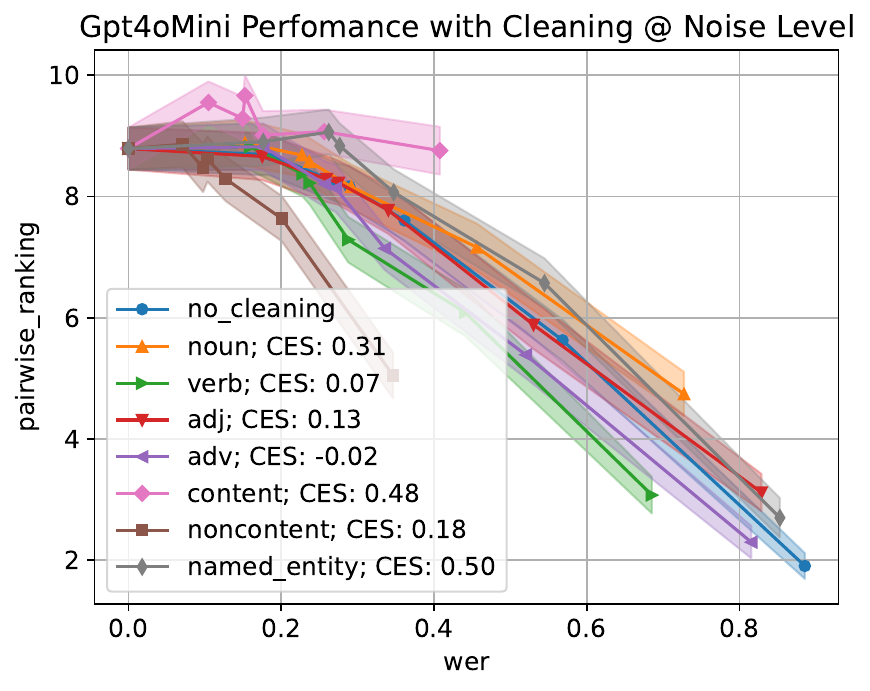}
    }
    \subfloat[ROUGE-1, GPT]{
        \includegraphics[width=0.32\linewidth, trim=8 8 7 21, clip]{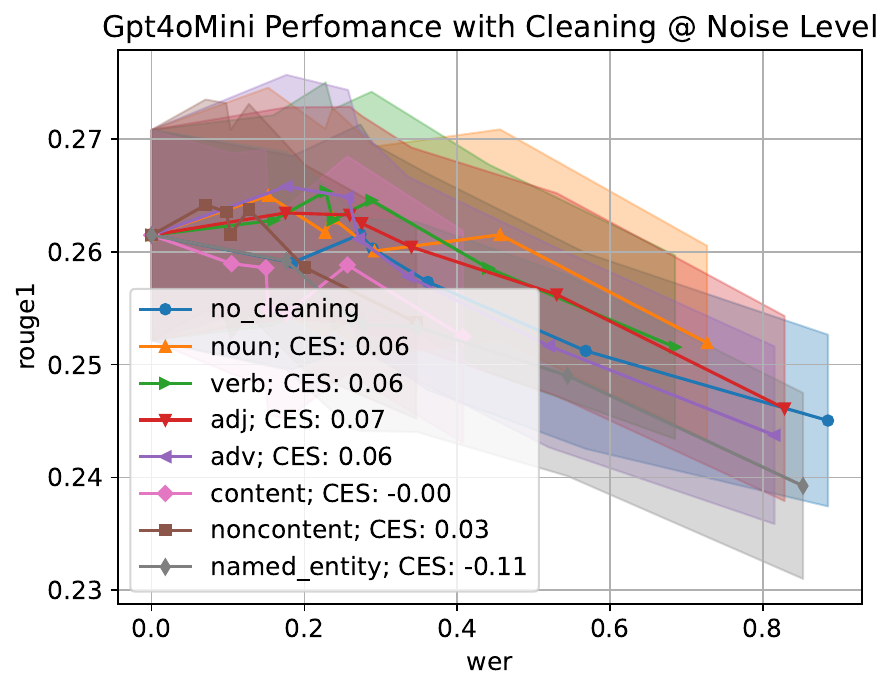}
    }
    \subfloat[ROUGE-2, GPT]{
        \includegraphics[width=0.32\linewidth, trim=8 8 7 21, clip]{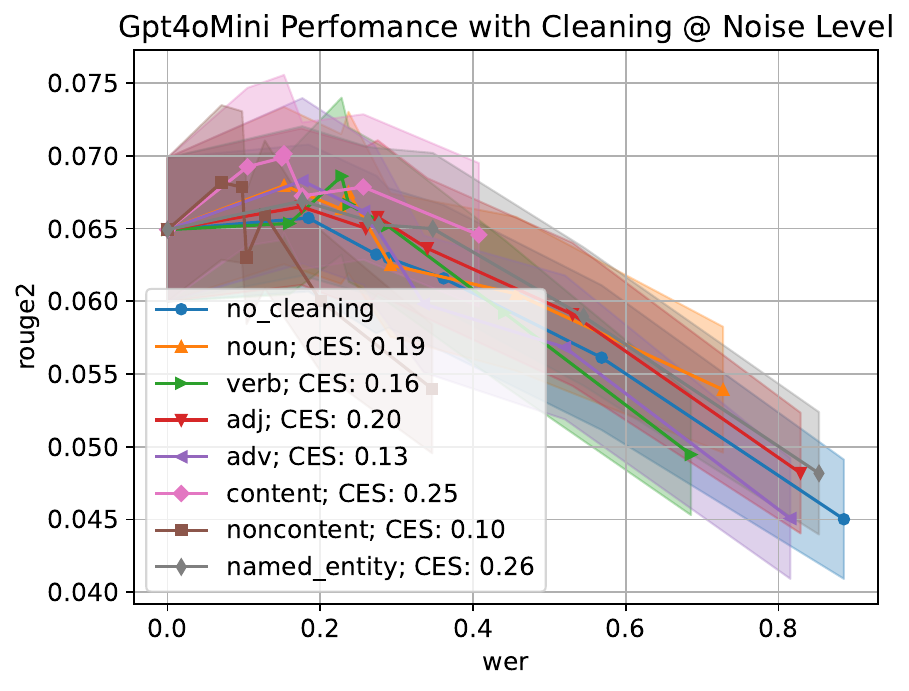}
    }
    \caption{The performance of models when applying various cleaning techniques, on the summarization dataset of \textbf{QMSum}. Each point on the ``no\_cleaning'' curve can be compared to the respective point on a cleaning technique's curve. A good cleaning technique should increase the task score (y value) as much as possible, with as little effort as possible (represented by decrease in WER, as the x value). Each cleaning technique is marked with its overall cleaning-effectiveness score which is computed as a function of the change in the task score and in the WER score. The CES scores can be seen also in \autoref{tab_scores_cleaning_all}.}
    \label{fig_cleaning_graphs_all_qmsum}
\end{figure*}

%% file: figures/cleaning/cleaning_main_all_qaconv.tex
\begin{figure*}[ht]
    \centering
    \subfloat[Fuzzy match, Mistral]{
        \includegraphics[width=0.32\linewidth, trim=8 8 7 21, clip]{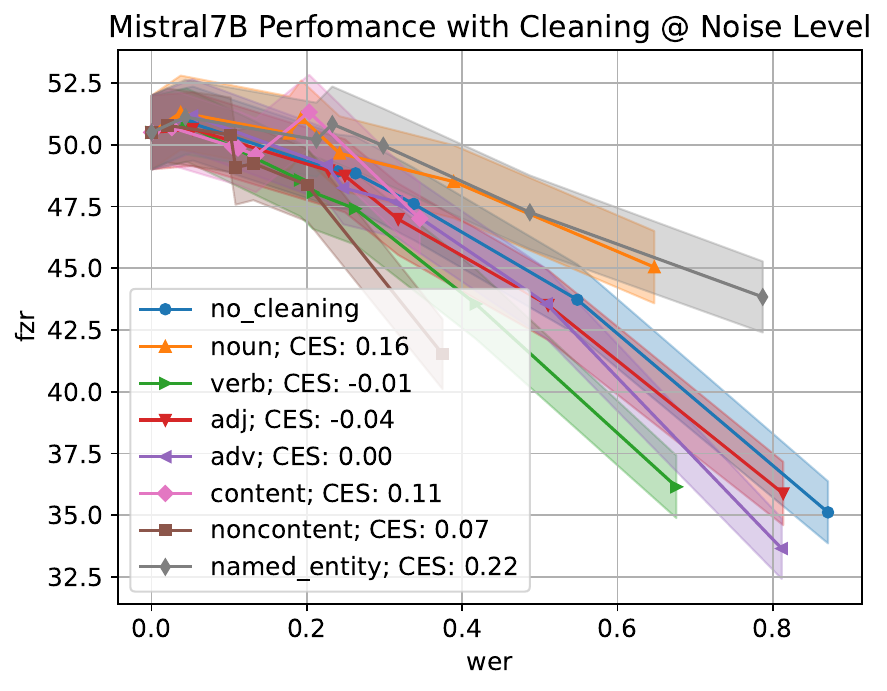}
    }
    \subfloat[Token $F_1$, Mistral]{
        \includegraphics[width=0.32\linewidth, trim=8 8 7 21, clip]{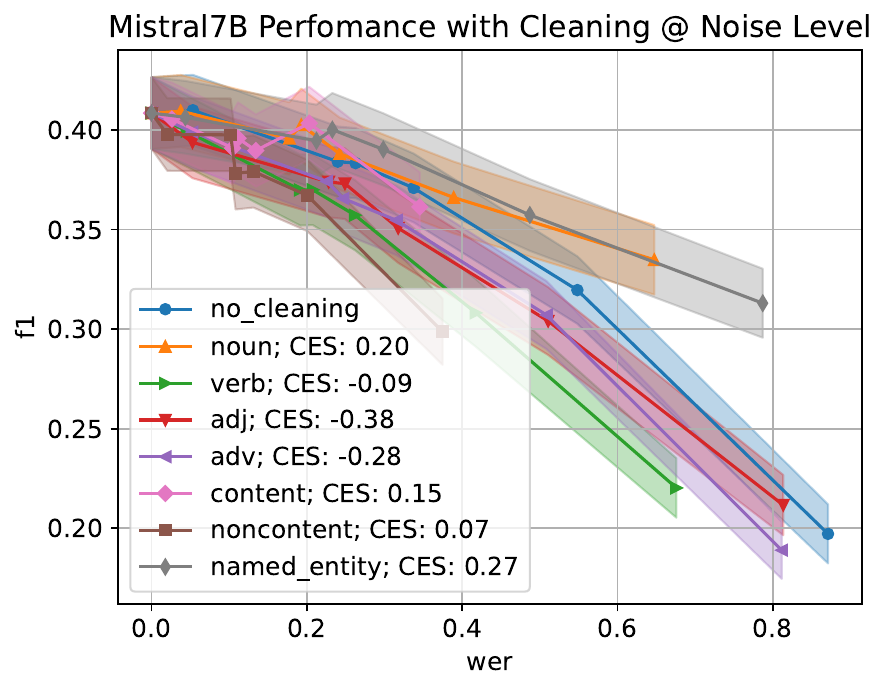}
    }
    \subfloat[Exact match, Mistral]{
        \includegraphics[width=0.32\linewidth, trim=8 8 7 21, clip]{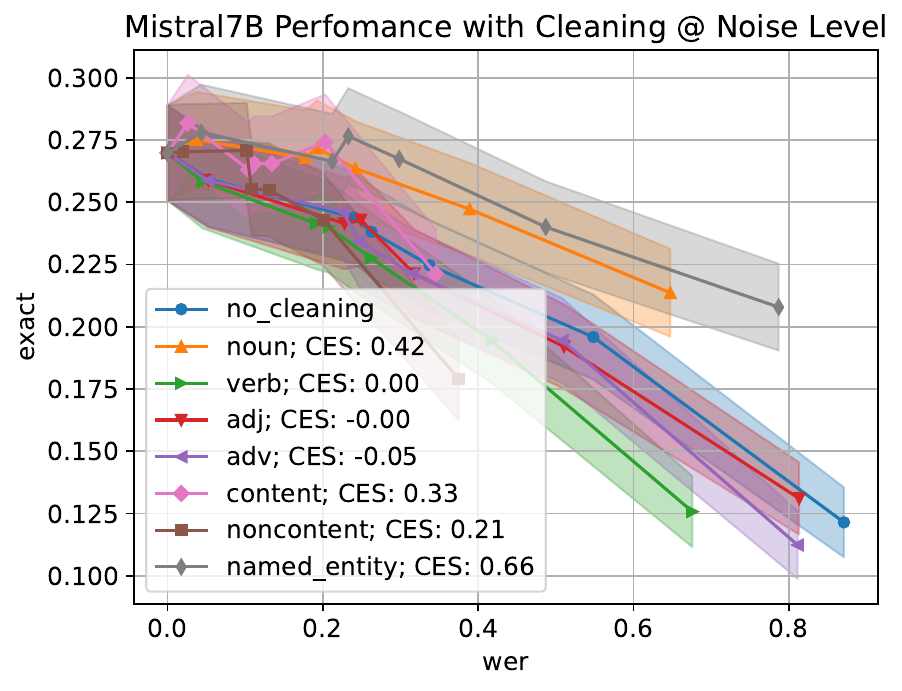}
    }
    \hspace{0.2cm}
    \subfloat[Fuzzy match, Llama3]{
        \includegraphics[width=0.32\linewidth, trim=8 8 7 21, clip]{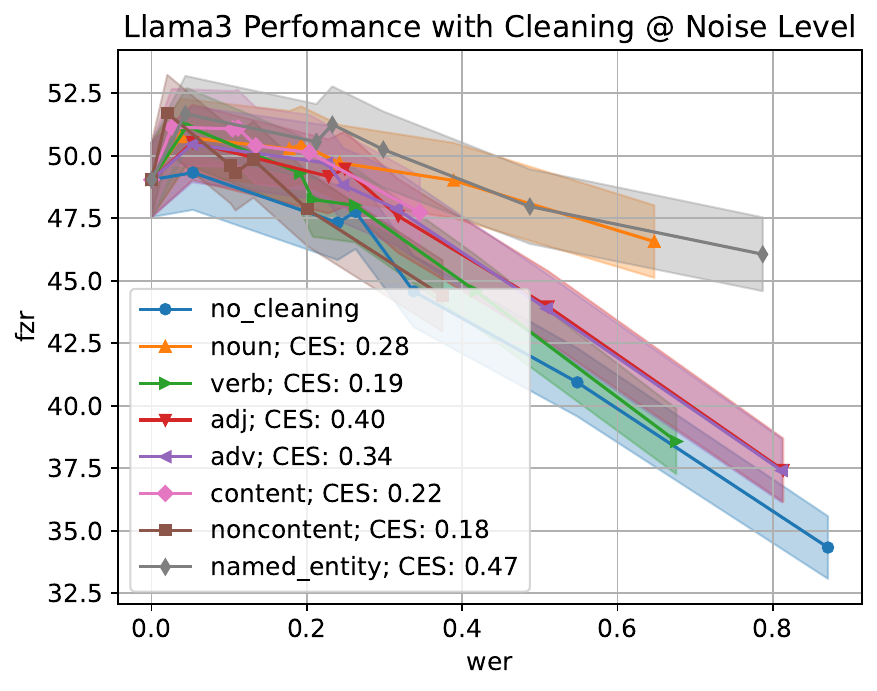}
    }
    \subfloat[Token $F_1$, Llama3]{
        \includegraphics[width=0.32\linewidth, trim=8 8 7 21, clip]{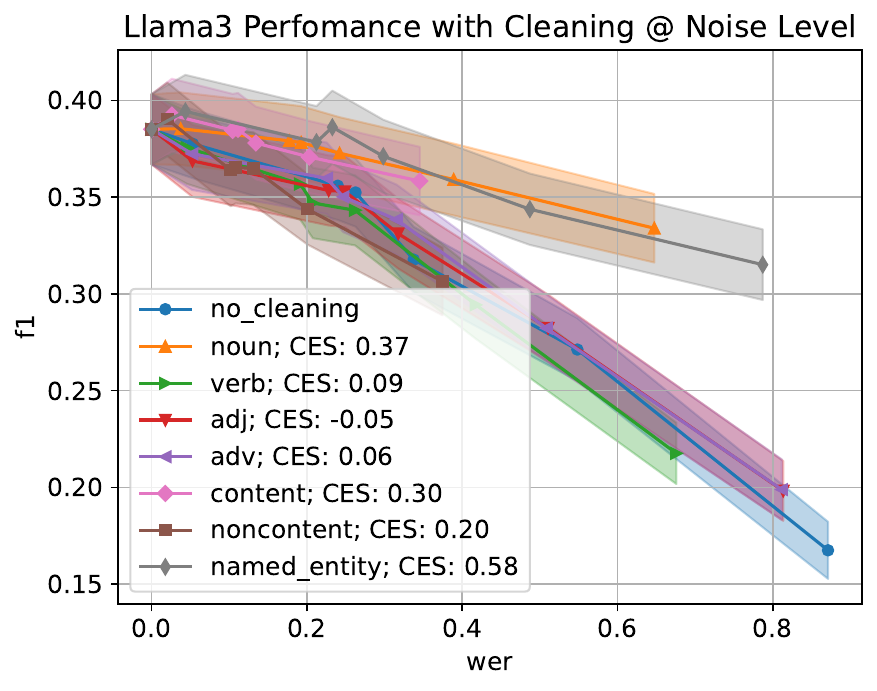}
    }
    \subfloat[Exact match, Llama3]{
        \includegraphics[width=0.32\linewidth, trim=8 8 7 21, clip]{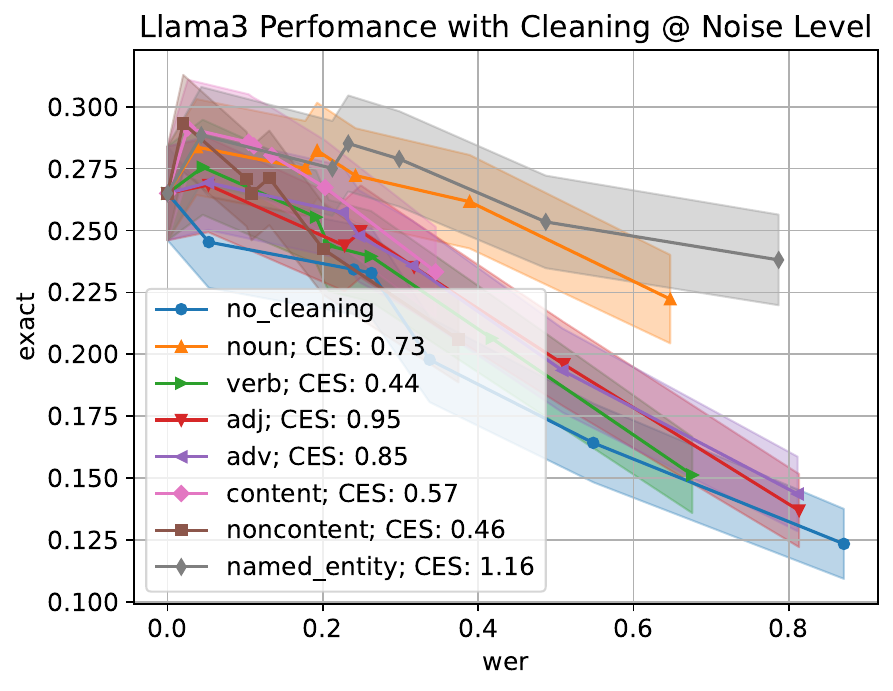}
    }
    \hspace{0.2cm}
    \subfloat[Fuzzy match, Llama3.1]{
        \includegraphics[width=0.32\linewidth, trim=8 8 7 21, clip]{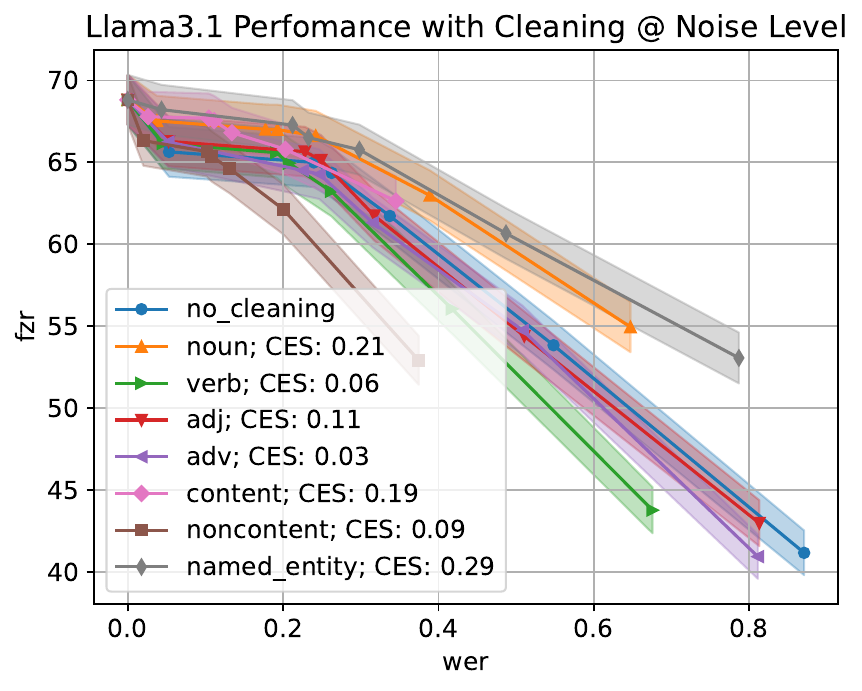}
    }
    \subfloat[Token $F_1$,  Llama3.1]{
        \includegraphics[width=0.32\linewidth, trim=8 8 7 21, clip]{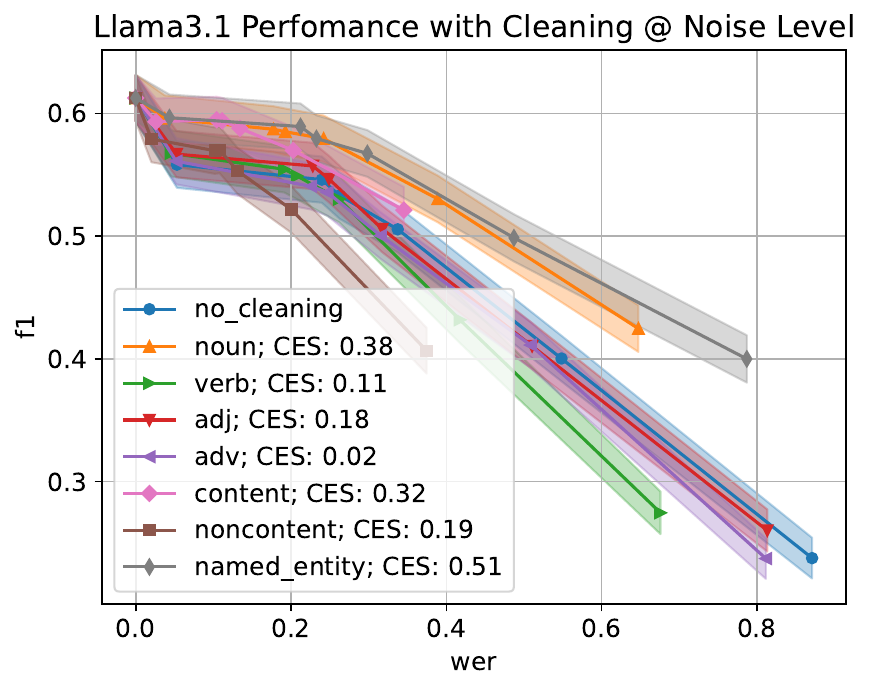}
    }
    \subfloat[Exact match,  Llama3.1]{
        \includegraphics[width=0.32\linewidth, trim=8 8 7 21, clip]{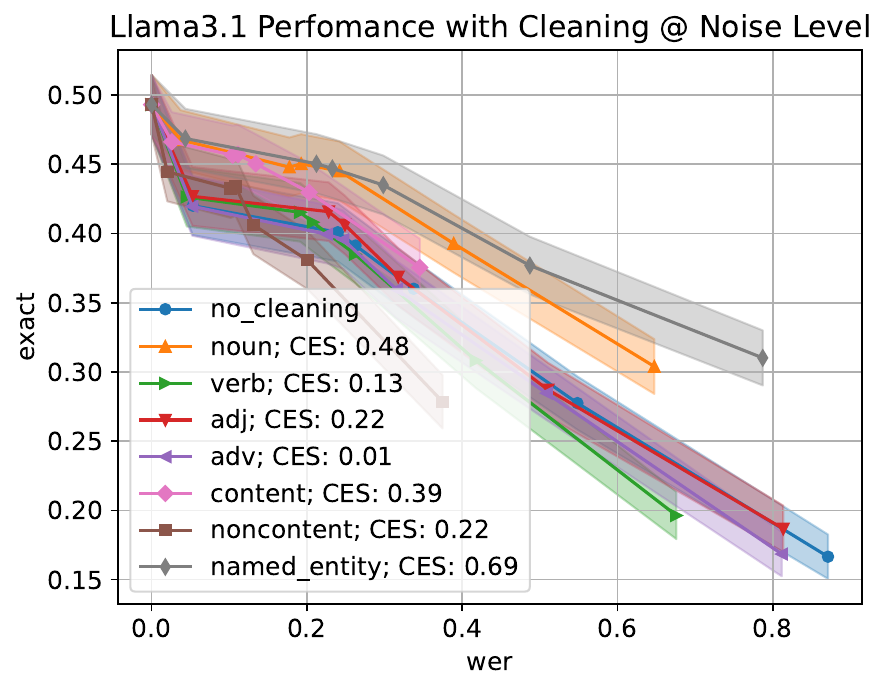}
    }
    \hspace{0.2cm}
    \subfloat[Fuzzy match, GPT]{
        \includegraphics[width=0.32\linewidth, trim=8 8 7 21, clip]{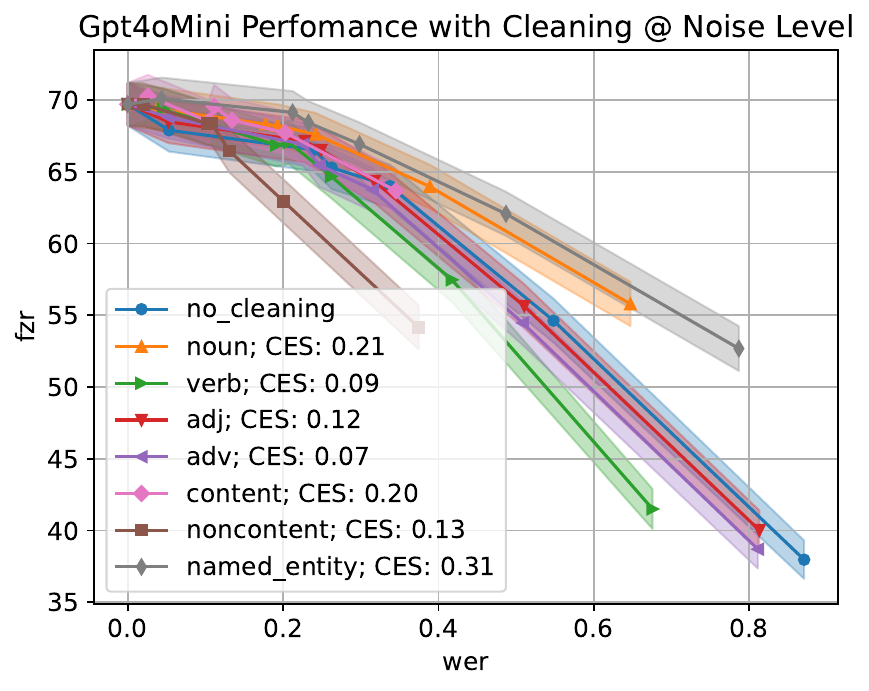}
    }
    \subfloat[Token $F_1$, GPT]{
        \includegraphics[width=0.32\linewidth, trim=8 8 7 21, clip]{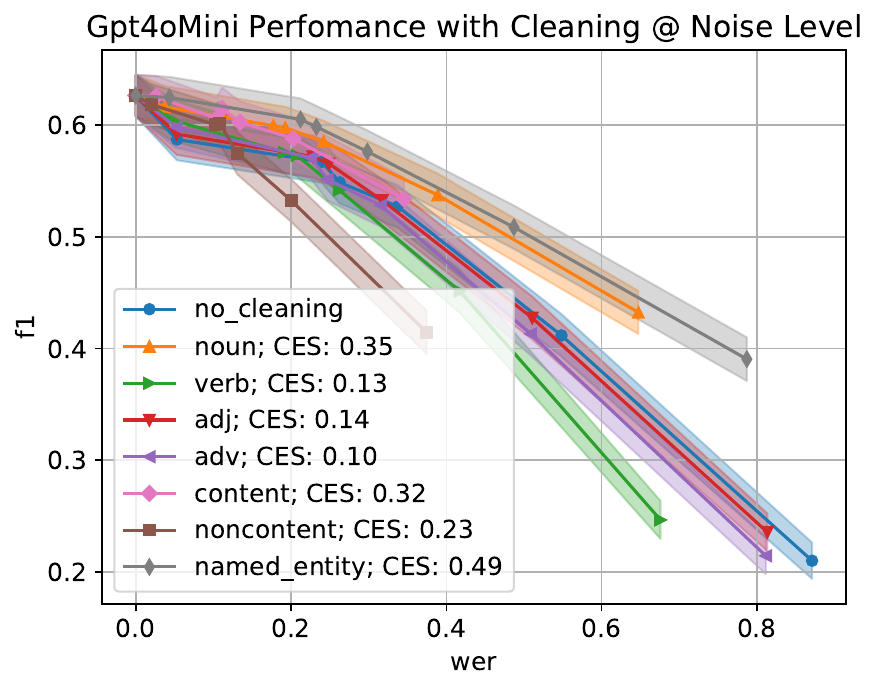}
    }
    \subfloat[Exact match, GPT]{
        \includegraphics[width=0.32\linewidth, trim=8 8 7 21, clip]{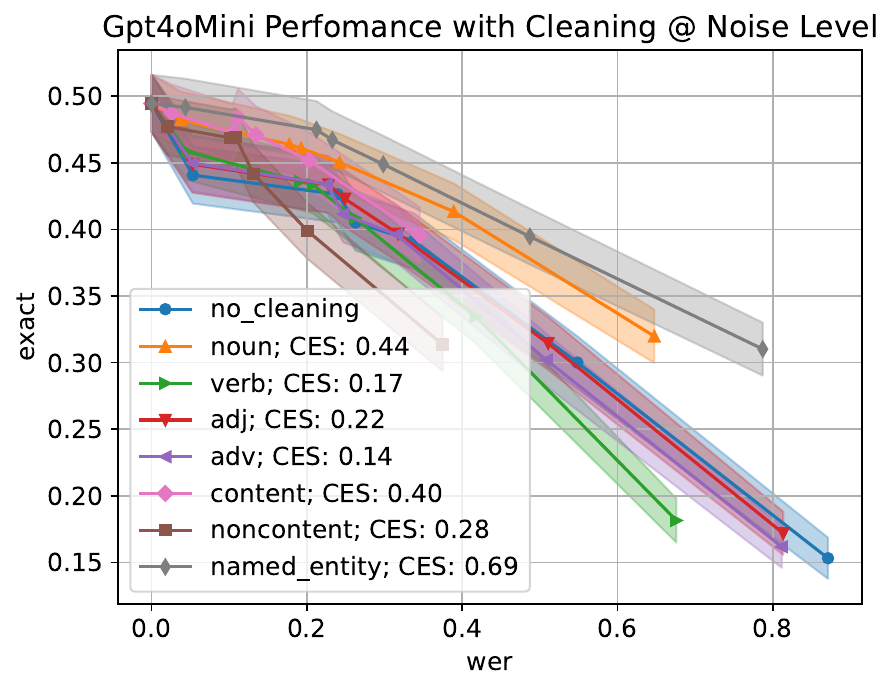}
    }
    \caption{The performance of models when applying various cleaning techniques, on the question-answering dataset of \textbf{QAConv}. Each point on the ``no\_cleaning'' curve can be compared to the respective point on a cleaning technique's curve. A good cleaning technique should increase the task score (y value) as much as possible, with as little effort as possible (represented by decrease in WER, as the x value). Each cleaning technique is marked with its overall cleaning-effectiveness score which is computed as a function of the change in the task score and in the WER score. The CES scores can be seen also in \autoref{tab_scores_cleaning_all}.}
    \label{fig_cleaning_graphs_all_qaconv}
\end{figure*}

%% file: figures/cleaning/cleaning_main_all_mrda.tex
\begin{figure*}[ht]
    \centering
    \subfloat[Macro-$F_1$, Mistral]{
        \includegraphics[width=0.32\linewidth, trim=8 8 7 21, clip]{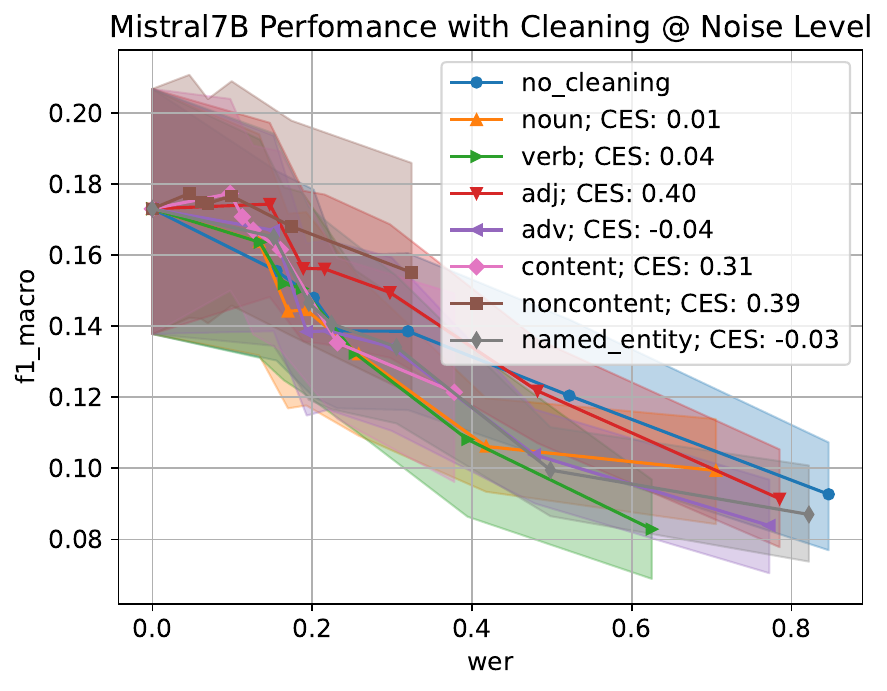}
    }
    \subfloat[Accuracy, Mistral]{
        \includegraphics[width=0.32\linewidth, trim=8 8 7 21, clip]{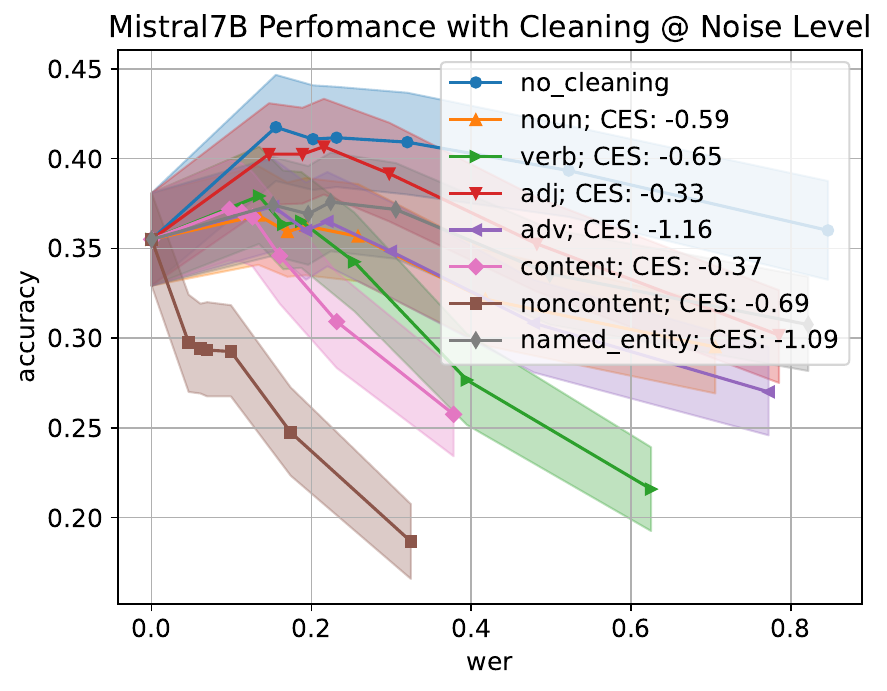}
    }
    \hspace{0.2cm}
    \subfloat[Macro-$F_1$, Llama3]{
        \includegraphics[width=0.32\linewidth, trim=8 8 7 21, clip]{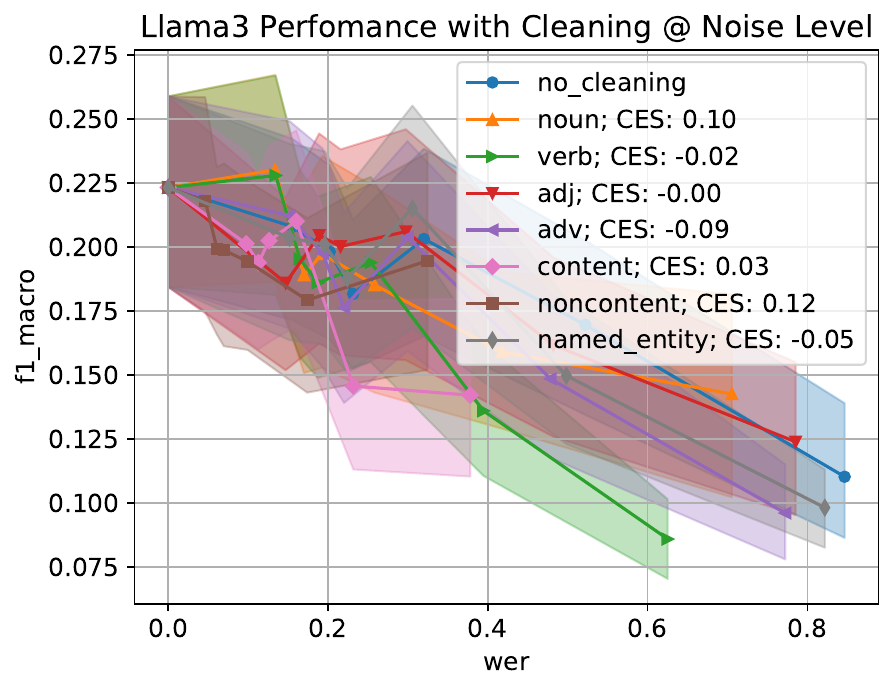}
    }
    \subfloat[Accuracy, Llama3]{
        \includegraphics[width=0.32\linewidth, trim=8 8 7 21, clip]{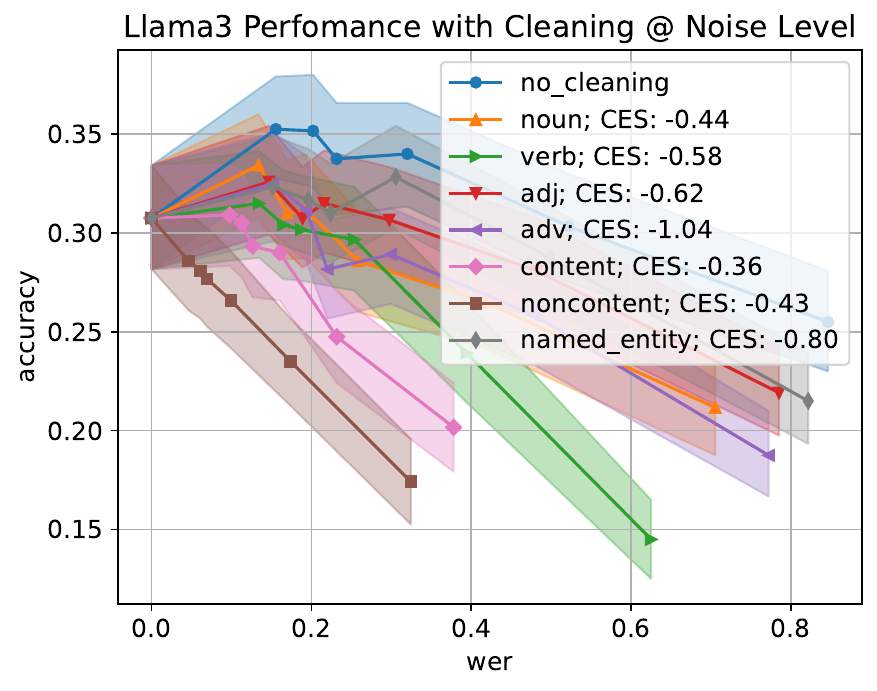}
    }
    \hspace{0.2cm}
    \subfloat[Macro-$F_1$, Llama3.1]{
        \includegraphics[width=0.32\linewidth, trim=8 8 7 21, clip]{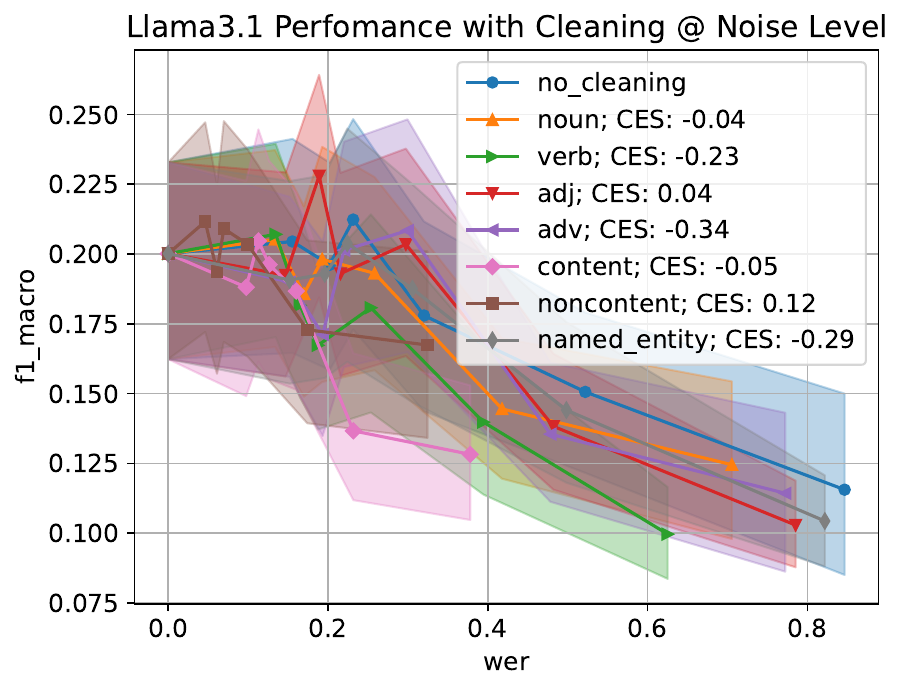}
    }
    \subfloat[Accuracy, Llama3.1]{
        \includegraphics[width=0.32\linewidth, trim=8 8 7 21, clip]{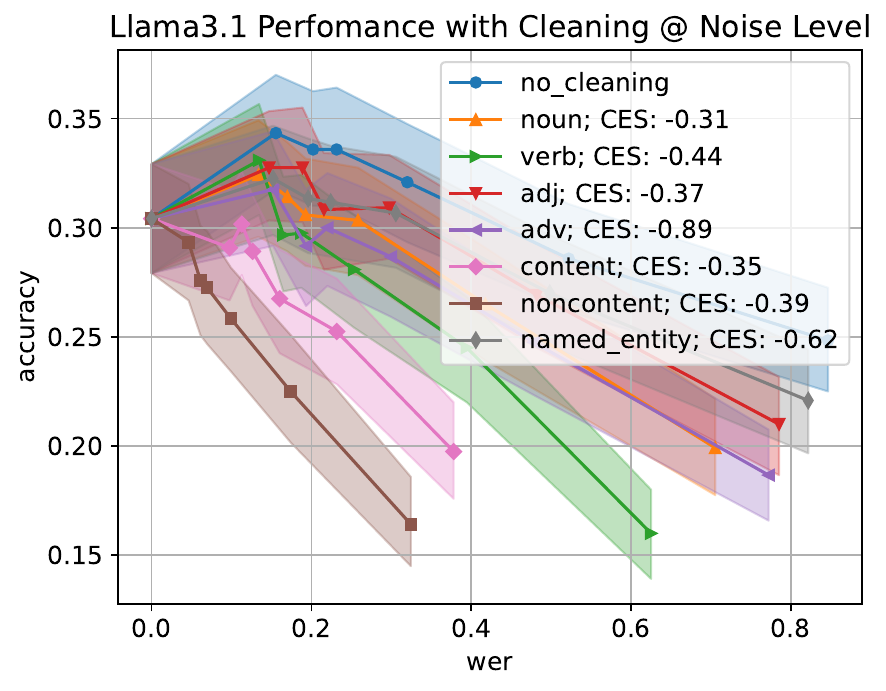}
    }
    \hspace{0.2cm}
    \subfloat[Macro-$F_1$, GPT]{
        \includegraphics[width=0.32\linewidth, trim=8 8 7 21, clip]{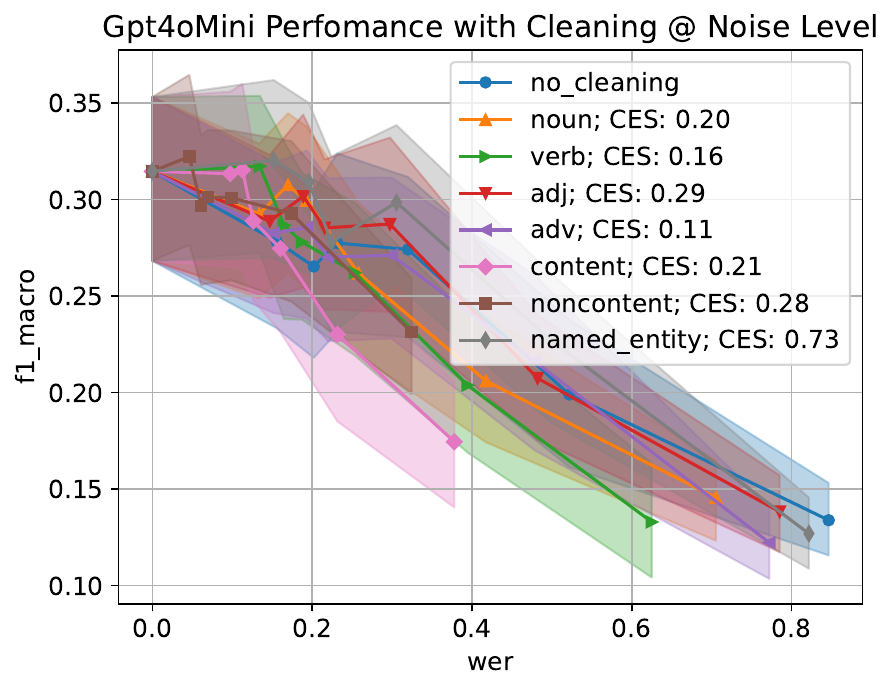}
    }
    \subfloat[Accuracy, GPT]{
        \includegraphics[width=0.32\linewidth, trim=8 8 7 21, clip]{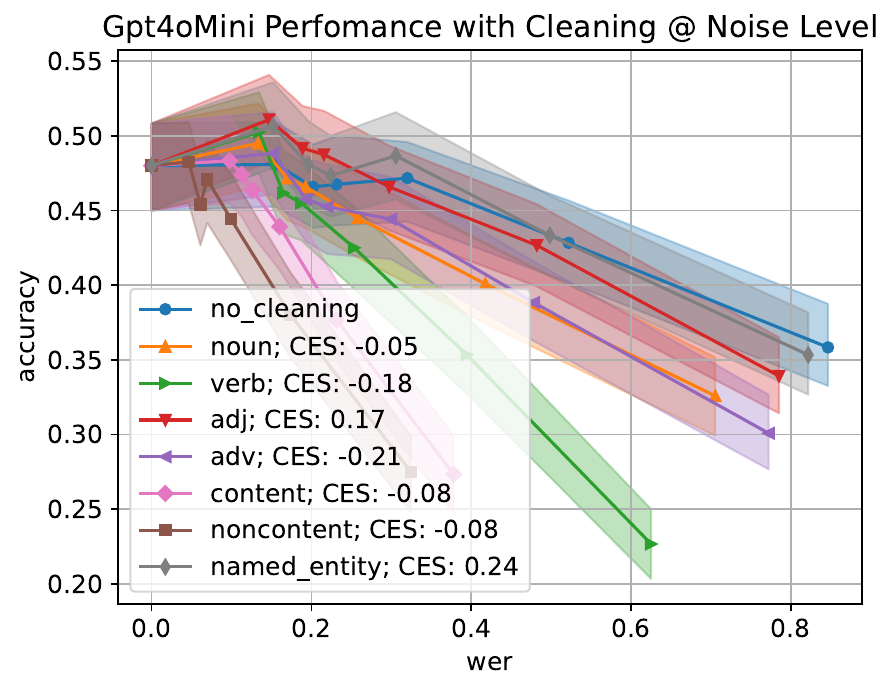}
    }
    \caption{The performance of models when applying various cleaning techniques, on the dialog-act classification dataset of \textbf{MRDA}. Each point on the ``no\_cleaning'' curve can be compared to the respective point on a cleaning technique's curve. A good cleaning technique should increase the task score (y value) as much as possible, with as little effort as possible (represented by decrease in WER, as the x value). Each cleaning technique is marked with its overall cleaning-effectiveness score which is computed as a function of the change in the task score and in the WER score. The CES scores can be seen also in \autoref{tab_scores_cleaning_all}.}
    \label{fig_cleaning_graphs_all_mrda}
\end{figure*}

%% file: figures/datasets/qmsum_example.tex
\begin{figure*}[t]
    \centering
    \begin{tcolorbox}[
      title=QMSum Example,
      colframe=blue!50!black,
      colback=blue!5!white,
      coltitle=white,
      fonttitle=\bfseries,
      boxrule=0.5mm,
      width=\textwidth,
      sharp corners
    ]

    \textbf{Instance Title:} education\_13 \\

    \textbf{Transcript:}
    \begin{itemize}
      \item \underline{Lynne Neagle AM}: Good afternoon, everyone. Welcome to the Children, Young People and Education Committee. We've received apologies...
      \item \underline{Barry Hughes}: Perfectly happy.
      \item \underline{Sian Gwenllian AM}: Thank you very much. I would like to start just by looking in general at how the law currently stands...
      \item ...
    \end{itemize}

    \vspace{0.5em}

    \textbf{Summaries:} 
    \begin{itemize}
        \item \textbf{Query:} Summarize the whole meeting. \textit{(Generic summary)}\\
        \textbf{Summary:} This meeting was the eleventh evidence session on the Children Abolition of Defense of Reasonable Publishment Wales Bill. Barry Hughes was...
        \item \textbf{Query:} Summarize the discussion about the efficacy of the law.\\
        \textbf{Summary:} Barry Hughes first stated that children had fewer rights than adults and therefore the law should be enforced to...
        \item ...
    \end{itemize}

    \end{tcolorbox}
    \caption{An example from the QMSum dataset. Each of the 35 transcripts has a generic summary and several query-focused summaries (avg. 8 summaries per transcript).}
    \label{fig_qmsum_example}
\end{figure*}

%% file: figures/datasets/qaconv_example.tex
\begin{figure*}[t]
    \centering
    \begin{tcolorbox}[
      title=QAConv Example,
      colframe=blue!50!black,
      colback=blue!5!white,
      coltitle=white,
      fonttitle=\bfseries,
      boxrule=0.5mm,
      width=\textwidth,
      sharp corners
    ]

    \textbf{Instance Title:} court-04-1506 \\

    \textbf{Transcript:}
    \begin{itemize}
      \item \underline{CHIEF JUSTICE ROBERTS}: We'll hear argument first this morning in 04-1506, Arkansas Department of Health and Human Services v. Ahlborn. Ms. Freno.
      \item \underline{MS. FRENO (PETITIONER)}: Mr. Chief Justice, and may it please the Court, The parties agree that Medicaid paid over \$215,000 to cover the costs of medical care...
      \item \underline{JUSTICE KENNEDY}: Can you tell me? It's my -- excuse me. My understanding was that Arkansas had intervened in the suit.
      \item ...
    \end{itemize}

    \vspace{0.5em}

    \textbf{QA:} 
    \begin{itemize}
        \item \textbf{Question:} What percentage of the Medicaid claim gets cut in half because of comparative negligence?\\
        \textbf{Answers:} 100 | 100 percent
        \item \textbf{Question:} How many cents will the Medicaid recipient get from the State?\\
        \textbf{Answer:} unanswerable
        \item \textbf{Question:} What would the State do if they wanted to pursue litigation?\\
        \textbf{Answers:} the case would just have to go forward to litigation | the case would just have to go forward | go forward
        \item ...
    \end{itemize}

    \end{tcolorbox}
    \caption{An example from the QAConv dataset, from which we only use the long spoken conversations. Each of the 505 transcripts has several questions (avg. 4.1) whose answers are extracted from the transcript. A question can have one or more answers (separated by a ``|'' in the figure), or it can be unanswerable.}
    \label{fig_qaconv_example}
\end{figure*}

%% file: figures/datasets/mrda_example.tex
\begin{figure*}[t]
    \centering
    \begin{tcolorbox}[
      title=MRDA Example,
      colframe=blue!50!black,
      colback=blue!5!white,
      coltitle=white,
      fonttitle=\bfseries,
      boxrule=0.5mm,
      width=\textwidth,
      sharp corners
    ]

    \textbf{Transcript Title:} Bmr022 \\

    \textbf{Utterances:}
    \begin{itemize} 
        \setlength\itemsep{0em}
        \item ...
        \item \underline{fe008}: well and then the other possibility was that we provide them with a a file that already has the beeps in it or something. \textbf{<Statement>}
        \item \underline{me011}: so \textbf{<Floor Holder>}
        \item \underline{me013}: huh. \textbf{<Continuer>}
        \item \underline{fe008}: wasn't that what you said? \textbf{<Yes-No-Question>}
        \item \underline{me018}:\\
        sort of a text template. \textbf{<Statement>}\\ 
        but i don't know how we can do that. \textbf{<Statement>}
        \item \underline{fe008}: uhhuh. \textbf{<Continuer>}
        \item \underline{me011}:\\
        yeah we don't know what their process is. \textbf{<Floor Grabber>}\\
        so i had two ideas. \textbf{<Statement>}\\
        the first was to provide them a text template that had both the beeps in it and the speaker i ds. \textbf{<Statement>}
        \item \underline{fe008}: yeah. \textbf{<Continuer>}
        \item \underline{me011}:\\
        you know just male female english nonenglish one two three four. \textbf{<Statement>}\\
        um \textbf{<Floor Holder>}
        \item \underline{me013}: can i i i just \textbf{<Yes-No-Question>}
        \item \underline{me011}: and they just filled it in. \textbf{<Statement>}
        \item \underline{me013}: how many how many beeps are there and how many do were they were were off? \textbf{<Wh-Question>}
        \item \underline{me011}:\\
        it was like a hundred twenty. \textbf{<Statement>}\\
        and they had a hundred twenty three or something like that. \textbf{<Statement>}
        \item \underline{me013}: all right. \textbf{<Statement>}
        \item ...
    \end{itemize}

    \end{tcolorbox}
    \caption{A snippet from a transcript in the MRDA dataset. Each utterance has at least one segment, and each segment (beginning with the underlined speaker ID) is labeled with its dialog act in bold. There are 12 transcripts, and we use a subset of 100 utterance segments per transcript for a total of 1200 instances.}
    \label{fig_mrda_example}
\end{figure*}